\documentclass{ieeeaccess}
\usepackage{cite}
\usepackage{amsmath,amssymb,amsfonts}
\usepackage{algorithmic}
\usepackage{textcomp}
\usepackage{hyperref}
\usepackage{rotating}
\usepackage{graphicx} 
\usepackage{rotating}
\usepackage{colortbl}
\usepackage{lscape}
\usepackage{longtable}
\usepackage{xtab}
\usepackage{amsmath}
\usepackage{epstopdf}
\usepackage[ruled,linesnumbered]{algorithm2e}
\usepackage{capt-of}
\usepackage{bussproofs}
\usepackage{siunitx}
\usepackage{multicol}
\usepackage{lipsum}
\usepackage{graphicx}
\usepackage{fixltx2e}
\usepackage{caption,setspace}
\usepackage[final]{pdfpages}
\usepackage{tablefootnote}
\usepackage{subcaption}
\usepackage{url}
\usepackage{breakurl}

\usepackage[table,xcdraw]{xcolor}
\usepackage{soul}

\usepackage[table,xcdraw]{xcolor}

\DeclareCaptionType{mycapequ}[][List of equations]
\def\BibTeX{{\rm B\kern-.05em{\sc i\kern-.025em b}\kern-.08em
		T\kern-.1667em\lower.7ex\hbox{E}\kern-.125emX}}

\begin{document}
	\history{}
	\doi{}	
	
	\title{Toward collision-free trajectory for autonomous and pilot-controlled unmanned aerial vehicles}
	\author{\uppercase{Kaya Kuru}\authorrefmark{1}
			\uppercase{John~Michael~Pinder}\authorrefmark{1}
			\uppercase{Benjamin~Jon~Watkinson}\authorrefmark{1}
			\uppercase{Darren~Ansell}\authorrefmark{1}
			\uppercase{Keith~Vinning}\authorrefmark{2}
			\uppercase{Lee~Moore}\authorrefmark{2}
			\uppercase{Chris~Gilbert}\authorrefmark{2}
			\uppercase{Aadithya~Sujit}\authorrefmark{1}
			\uppercase{and~David~Jones}\authorrefmark{1}				
	}	
	\address[1]{School of Engineering, University of Central Lancashire, Preston, PR1 2HE UK}
	\address[2]{PilotAware Ltd, 19 John McGuire Crescent, Coventry, CV3 2Q, UK (https://www.pilotaware.com)}

	\markboth
	{Kuru \headeretal: Towards mid-air collision-free trajectory for autonomous and pilot-controlled unmanned aerial vehicles}
	{Kuru \headeretal: Towards mid-air collision-free trajectory for autonomous and pilot-controlled unmanned aerial vehicles}
	
	\corresp{Corresponding author: Kaya Kuru (e-mail: kkuru@uclan.ac.uk).} 
	
	\begin{abstract}		
		The subject of autonomy within unmanned aerial vehicles (UAVs) has proven to be a remarkable research field -- mostly due to the development of AI techniques within embedded advanced bespoke microcontrollers -- during the last several decades. For drones, as safety-critical systems, there is an increasing need for onboard detect \& avoid (DAA) technology i) to see, sense or detect conflicting traffic or imminent non-cooperative threats due to their high mobility with multiple degrees of freedom and the complexity of deployed unstructured environments, and subsequently ii) to take the appropriate actions to avoid collisions depending upon the level of autonomy. The safe and efficient integration of UAV traffic management (UTM) systems with air traffic management (ATM) systems, using intelligent autonomous approaches, is an emerging requirement where the number of diverse UAV applications is increasing on a large scale in dense air traffic environments for completing swarms of multiple complex missions flexibly and simultaneously. Significant progress over the past few years has been made in detecting UAVs present in aerospace, identifying them, and determining their existing flight path. This study makes greater use of electronic conspicuity (EC) information made available by PilotAware Ltd in developing an advanced collision management methodology -- Drone Aware Collision Management (DACM) -- capable of determining and executing a variety of time-optimal evasive collision avoidance (CA) manoeuvres using a reactive geometric conflict detection and resolution (CDR) technique. The merits of the DACM methodology have been demonstrated through extensive simulations and real-world field tests in avoiding mid-air collisions (MAC) between UAVs and manned aeroplanes. The results show that the proposed methodology can be employed successfully in avoiding collisions while limiting the deviation from the original trajectory in highly dynamic aerospace without requiring sophisticated sensors and prior training. With the proposed technological improvement equipped with Artificial Intelligence (AI) techniques, MAC risks which cannot be avoided with the current off-the-shelf sensor technologies, in particular, between flights with very high velocities, can be definitely prevented with the accurate measurements and state and situation awareness (SSA) that uses a global coverage strategy with real-time low latency EC data feeds acquired from all aircraft. The MAC standards, dictated by the aviation authorities, can be mandated for UAVs considering the reliable decision-making abilities of DACM -- without creating new collision risks during evasive manoeuvres, which can expedite the safe and efficient integration of UAVs into ATM systems.		
	\end{abstract}
	
	
	\titlepgskip=-15pt
	
	\maketitle
	\section{Introduction}
	\label{sec:introduction}
\IEEEPARstart{T}{he} exponential growth of interest and research in UAVs is strongly pushing for the emergence of autonomous flying robots~\cite{4337970}. Drone Industry Insights (DRONEII) categorises the autonomy with 5 levels~\cite{MRadovic19} based on degrees of independence, namely, 1: low automation (i.e., the UAV has control of at least one vital function, with a pilot in control); 2: partial automation (i.e., the UAV can take over heading and altitude under certain conditions with a pilot still responsible for safe operation); 3: conditional automation (i.e., the UAV can perform all functions and a pilot act as a fall-back system); 4: high automation (i.e., the UAV has back-up systems, so if one fails, the platform is still operational and a pilot is out of the loop); 5: full automation (i.e, the UAV can plan its actions using advanced AI autonomous learning techniques) with little or no human intervention in the control loop. As the level of autonomy increases, UAVs can operate in more complex environments and execute more complex tasks with less prior knowledge and fewer operator interactions~\cite{doi:10.1002/tee.23041}.

The game-changing role of UAVs relies on their safe use, particularly, in an autonomous manner with no threat to the public and other aircraft by using effective collision management (CM) approaches. Unlike autonomous ground vehicles (AGVs), self-operating UAVs with six degrees of freedom (DoF) have a higher dimensional configuration space, which makes the motion planning of multi-UAVs a challenging task concerning the other nearby flights; in addition, uncertainties and noises are more significant in UAV scenarios, which increases the difficulty of autonomous navigation without colliding where reliable situation awareness (SA) is difficult to achieve in real-world applications due to the imperfect sensing~\cite{9001167}. The ability to navigate autonomously is the most critical component of UAV automation both to complete the mission and to reach the targeted destination safely and efficiently by avoiding unexpected encounters in the trajectory. Coordinated task assignment is a key scientific issue for autonomous control of UAVs~\cite{9847033}. As swarm intelligence research advances, multiple drones are likely to be used increasingly in industrial sectors, sometimes, in achieving specific coordinated tasks, carrying a heavy payload together with a robotic arm or a suspended cable grasped by a fixed gripper, search and rescue by assigning different target region of interest (RoI) to multiple A-UAVs. When using swarms of drones, the problem of mid-air collisions (MAC) between UAVs and manned aeroplanes presents a serious safety risk. Furthermore, a small size combined with the relatively high speed of UAVs makes their detection via a pilot's visual capabilities very difficult~\cite{9256553}. The rapidly increasing number of UAVs, either in autonomous mode or pilot-controlled mode poses a security challenge with the imminent traffic chaos that needs to be addressed urgently~\cite{9314128}. The lack of appropriate real-time decision-making strategies is the cause of many accidents involving UAVs according to the reports presented by the US Federal Aviation Administration~\cite{8379533}. Reactive artificial potential field (APF)~\cite{8573148}, vision/sonar-based (Section~\ref{sec:Literature})) may be effective for low-traffic scenarios where a drone needs to avoid obstacles or aircraft using onboard proximity sensors. However, as air traffic volumes inevitably increase, these techniques will become less effective. More specifically, onboard detect \& avoid (DAA) technology is particularly challenging in high aerial traffic scenarios due to the processing of large volumes of data at high speeds, which is required to provide real-time optimised avoidance strategies. A reactive collision avoidance (CA) manoeuvre at close-range may result in an action that successfully avoids one drone or aircraft, but could potentially move the drone into the collision course of another. Thus creating a chain reaction of collisions resulting from improper planning and prediction. Continuous real-time positional awareness of every drone or aeroplane will allow for more effective planning and prediction of suitable CM strategies by maintaining adequate separation enabling higher volumes of safe operation. There must be a robust system to allow the high volumes of aerial traffic to co-exist safely and where possible, respond without the need for human intervention for keeping each UAV well clear of other traffic using effective DAA techniques. This response requires each UAV to be aware of the position and trajectory of every other close-range flight. In addition, more advanced methods of CA must be employed to ensure the safe passage of multiple aircraft. Significant progress over the past few years has been made in detecting flights present in aerospace, identifying them, and determining their existing flight path. The innovative aspect of this paper involves building upon tried and tested commercially available PilotAware system to provide comprehensive inputs into a new advanced CM methodology, the so-called DroneAware CM (DACM) system. Recent advances in cyber-physical systems (CPS) within the concepts of Internet of Everything (IoE) and Automation of Everything (AoE)~\cite{8675275} allows us to teleoperate remote objects using Digital Twins (DTs). In this respect, the DACM system was developed to create the DTs of aerial traffic (Fig.~\ref{fig:MainComponents}) by tightly communicating with the decentralised PilotAware ATOM-GRID Network. To clarify the novelty of this paper, particular contributions are outlined as follows.

1) DACM, equipped with a novel CM methodology within an autonomous control framework using geometry formation of flights, is built to enable Beyond Visual Line of Sight (BVLOS) operations with collision-free trajectories.

2) The system, primarily, aiming to address the mid-air collision (MAC), can perceive the surrounding dynamic aerospace environment precisely using a global coverage strategy with real-time low latency EC data feeds acquired from all aircraft and react efficiently to multiple nonlinear collision risks at a time with minimum trajectory deviations requiring no sophisticated sensors.

3) DACM in both air only or air/ground modes not only implements time-optimal CA for fully autonomous UAVs (FA-UAVs), but also manages pilot-controlled UAVs (PC-UAVs) in improving operational safety since it might be difficult for pilots to detect surrounding close-range/high-speed flights.

In the rest of the paper, the related works are investigated in Section~\ref{sec:Literature}. The proposed system is revealed in Section~\ref{sec:Methodology}. The experimental design is delineated in Section~\ref{sec:Design} and the results are presented in Section~\ref{sec:Results}. Discussions are provided in Section~\ref{sec:discussion}. Finally, Section~\ref{sec:conclusion} concludes with key findings and future works followed by limitations. 

\section {Related Works}
\label{sec:Literature}
Laser/sonar/radar (e.g.,~\cite{ViqueratAD2008RCAf},~\cite{8396612}) and vision-based sensors with lightweight and low-cost cameras (e.g.,~\cite{8684792}) are mainly deployed in CA systems developed using various Machine Learning (ML), AI (e.g., Deep Reinforcement Learning (RL)~\cite{9001167}) techniques. Vision-based navigation with providing abundant online information of the surroundings (e.g., colour, texture, and other visual information) and anti-interference ability proves to be a primary and promising research direction of autonomous navigation with the rapid development of computer vision by means of image processing~\cite{doi:10.1080/10095020.2017.1420509}. A vision-only-based CA approach for UAVs was proposed in~\cite{7535138} without using distance information. CA systems using distance-limited ranged vision-only based sensors are highly susceptible to harsh weather conditions due to the perception abilities of traditional vision sensors. Additionally, early detection of flights with high speed is difficult using those sensors to be able to determine and react properly in a timely manner. An even bigger problem with these sensors is that it is highly difficult to measure the required accurate distance/depth using cameras alone despite the recent advances in monocular cameras using visual odometry cues, stereo cameras using the parallax principle and RGB-D cameras using infrared sensors. On the other hand, laser/sonar/radar-based systems, with less susceptible to harsh weather conditions and the ability to measure the depth and velocity accurately, are employed to reveal instant SA leading to advanced CA, but with other bottlenecks such as increasing cost and requiring more payload capacity~\cite{8611082}. Besides, they are incapable of getting enough information in the complex environment due to their limited field of view and measurement range~\cite{doi:10.1080/10095020.2017.1420509}. The usage of a single sensor in UAV CA systems leads to a couple of deficiencies. To this end, various onboard complementary sensors are employed for reliable three-dimensional (3D) CA using more accurate SA acquired from the fusion of multiple sensors. However, complementary sensor technologies may exacerbate the limited payload capacity and battery life requiring more processing for fusing multiple sensor data. Furthermore, these types of CA techniques, with high computational time, cannot be scaled to hundreds of UAV agents at a time~\cite{8767930}.

For UAV avoidance, the geometric guidance algorithms are potentially best suited since they do not need to conduct extensive predictions and analyses without requiring large computational resources on-board~\cite{2015JGCD...38.1140J}. In the literature, there have been few works focusing on the area of dynamic obstacle (i.e., moving object) avoidance for CA~\cite{9145644}, in particular, based on UAV geometry formation as discussed in this paper. Jenie et al.~\cite{doi:10.2514/1.G001715} analysed the CA system of UAVs regarding aeroplanes by changing the vehicle velocity vector based on the encounter geometry using the velocity projection. Jenie et al.~\cite{2015JGCD...38.1140J} developed the selective velocity obstacle method (SVO) as an extension of the former velocity obstacle method. Xiuxia et al.~\cite{8536788} propose a reactive geometric 3D avoidance manoeuvre direction for a UAV to resolve conflicts among the dynamic objects using the collision-cone approach. The CA strategies for multiple UAVs using geometry conflict are investigated in~\cite{7947166} by expanding the collision-cone approach. Yang et al.~\cite{doi:10.2514/1.G002607} implemented a geometry-based distributed cooperative conflict resolution method to autonomously handle multiple encounters in integrated aerospace. Lindqvist et al.~\cite{9145644} propose a nonlinear model for predictive control, navigation and obstacle avoidance by predicting future positions. 

Most of the literature studies rely on onboard sensors to avoid any collision. However, the space for avoidance will be limited by the sensors with the range of detection with the UAV six DoF and a resolution manoeuvre has to be aggressively conducted with the maximum performance of UAVs by comprehending several encounters in the air traffic at once to be able to achieve safe flight as fast as possible~\cite{8536788}. The limited autonomous navigation capability severely hampers the application of UAVs in complex environments~\cite{doi:10.1080/10095020.2017.1420509}. None of the aforementioned near-optimal solutions has been found to overcome the mid-air collision-free autonomous UAV navigation, in particular, they have serious shortcomings in satisfying the MAC avoidance standards which are the main criteria to integrate UAVs into ATM systems concerning safety with UAV BVLOS operations. An effective optimal CA methodology that can be implemented in planning UAV local trajectory regarding collision risks is in high demand where the surrounding dynamic UAV environment is highly unpredictable concerning the very high speed of flights. This leads to high collision risks before taking appropriate action regarding the relative separation from the flight in collision risk and creation of new collision risks with the other flights in the separation manoeuvring direction, while a manoeuvring UAV is unaware of the other close range flights. It is worth noting that as part of the development and evaluation of collision avoidance systems between small and slow drones, the near MAC (NMAC) borders are defined as a cylindrical boundary where the vertical separation is less than 100 ft and the horizontal separation is less than 500 ft~\cite{lebron1983system}. Well-clear for these small and slow drones is defined by the UAS Executive Committee (EXCOM) Science and Research Panel (SARP) as simultaneously lost of horizontal separation of 2000 ft and vertical separation of 250 ft ~\cite{doi:10.2514/1.D0091}.
Close encounters between UAVs and manned aeroplanes are defined as when a MAC is likely within a 30-60 second time window~\cite{doi:10.2514/1.D0091} for a reasonable evasive manoeuvre to avoid a MAC. 1 MAC corresponds to around every 10 NMACs~\cite{doi:10.2514/6.2010-9333}. Regarding these definitions, it would be reasonable to assume that all UAVs, with the current sensor technologies concerning their limitations, are unaware of each other and other manned aeroplanes during mid-air encounters considering the combined speeds of two aircraft (e.g., black circles in Fig.\ref{fig:Deviation}). In this direction, it can be concluded that UAVs cannot be safely integrated with ATM systems considering these standards and the limitations. Complete and reliable SA about surrounding traffic is a key prerequisite of safe operations in any aerospace~\cite{9256553}.

Low-power EC systems (e.g., FLARM, ADS-B~\cite{9256553}) are already commercially available however, they provide traffic information; whilst minimising the impacts on payload capacity and radio frequency spectrum congestion, they must be integrated with self-separation and CA algorithms to fulfil the functionality of a complete DAA solution~\cite{pointon2018integration}. They are harnessed for providing effective traffic surveillance functions (e.g., conflict management by the pilot based on SA, air traffic monitoring, recommending early escape manoeuvrers to manned aeroplanes to avoid conflicts using TCAS, ACAS systems~\cite{USDoT11}), particularly for manned aeroplanes. These systems have not been certified as an acceptable means of compliance with UAS DAA requirements where manned aircraft performance assumptions do not apply to UASs~\cite{doi:10.2514/1.D0091}, which necessitates the development of new UAS DAA technologies. The proposed mechanism of incorporating EC devices into the operation of unmanned aircraft systems (UAS) was tested at an airfield in the UK in 2017 to measure their detection signature in km~\cite{8396612}. NASA Langley Research Center conducted a series of flight tests
to demonstrate the use of onboard autonomy-enabling technologies in scenarios where a non-conforming UAV flies through the assigned aerospace of another vehicle while trying to reach an emergency landing site~\cite{9256776}; in these highly restricted flight tests using FLARM EC, a scripted emergency scenario is triggered to make an emergency landing to avoid the collision.

To the best of our knowledge, based on the extensive literature survey, this research is the first comprehensive work that integrates fully autonomous UAVs with EC devices enabling an effective CM mechanism that leads to safe autonomous BVLOS operations by taking the remote operators out of the loop not only within UTM systems, but also within the integrated aerospace of UTM and ATM systems where the chance of a drone colliding with an aircraft has increased substantially as acknowledged by authorities (e.g.,~\cite{IFGreen18}). And again, to the best of our knowledge, EC, with an intelligent CM system, has been tested by incorporating UAVs and manned aircraft into the same aerospace for the first time to address mid-air collision. This paper contributes to the literature by proposing a novel 3D dynamic CM methodology, which has several advantages over other methods by handling multiple collision risks with no sophisticated sensors and no prior training. The PilotAware system is unique in the fact that it is capable of detecting more UAVs/aircraft types than any other system currently available on the market (detailed in Section~\ref{sec:Background}) which is a considerable benefit for the proposed CA methodology in this paper. With the rich set of data available, the developed DACM system is capable of performing exceptionally well at solving the challenge of DAA by considering the highest proportion of monitored flights to inform its data-driven decision-making capabilities. More specifically, collision risk assessment and modelling of CA for air vehicles are investigated in this study.

\section {Methodology}
\label{sec:Methodology}
\begin{figure}[]
	\begin{center}
		\includegraphics[width=0.43\textwidth]{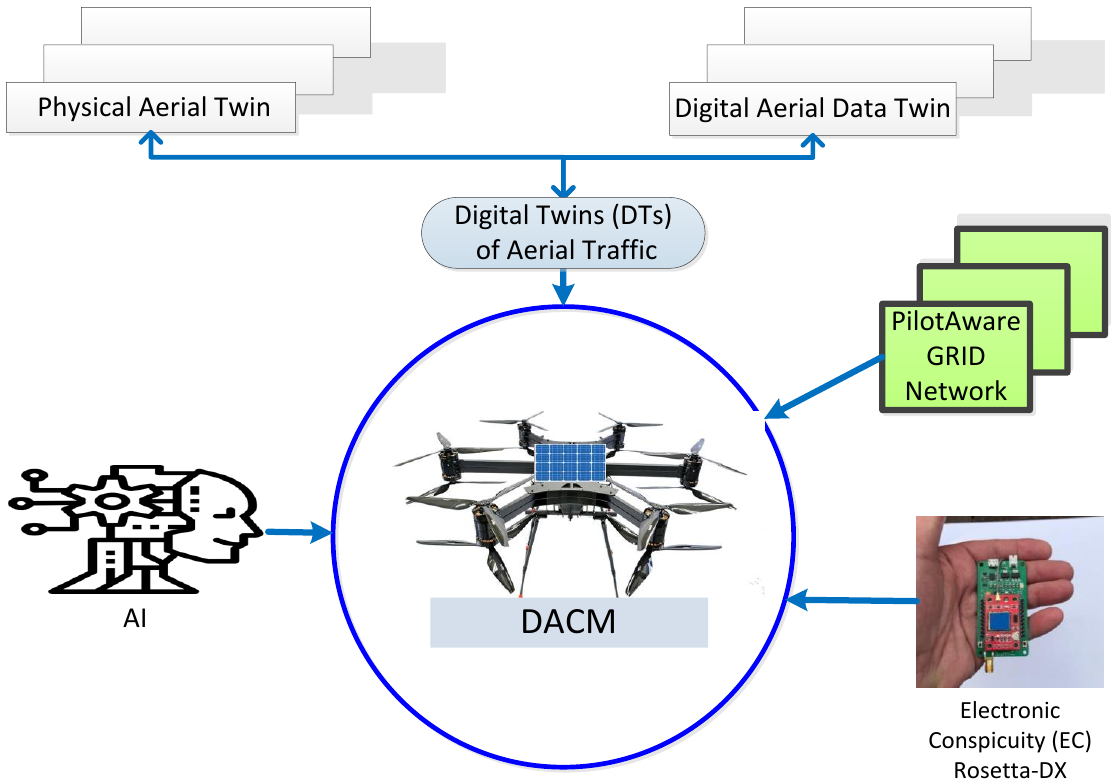}
	\end{center}
	\caption{Main building blocks of the DACM system.}
	\label{fig:MainComponents}
\end{figure}

\begin{figure*}[htp]
	\begin{center}
		\includegraphics[width=\textwidth]{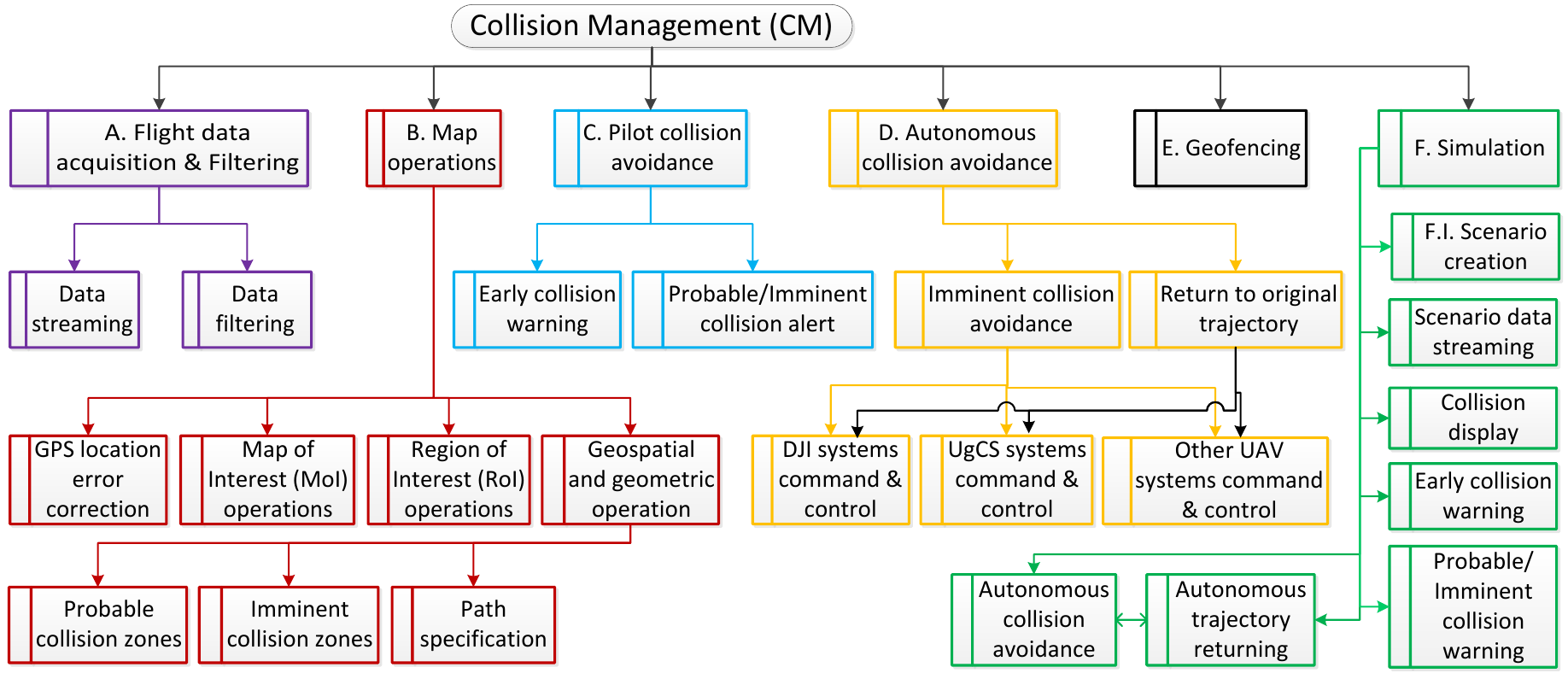}
	\end{center}
	\caption{Main modules and their submodules in the system.}
	\label{fig:modules}
\end{figure*}

Readers are referred to a YouTube video~\footnote{https://www.youtube.com/watch?v=Lr9yI3Hlfp8} for the summary of the research. An effective and efficient CM system, the so-called DACM system, is developed for FA-UAVs and PC-UAVs involving semi-autonomous UAVs (SA-UAVs). An artificial intelligence (AI) technique uses path planning to calculate trajectories of other UAVs/aircraft and predicts a probabilistic navigation path to generate an optimised avoidance strategy whilst complying with the pre-configured level of autonomy. The components of the proposed system are presented in Fig.~\ref{fig:MainComponents}. These components are elaborated in the following subsections. More specifically, the modules of DACM are introduced in Fig.~\ref{fig:modules} and the methodology using these modules is depicted in Fig.~\ref{fig:methodology}. The background of the study is provided in Sections~\ref{sec:Background} before revealing the developed techniques in the system in~\ref{sec:Techniques}.

\begin{figure*}[htp]
	\begin{center}
		\includegraphics[width=\textwidth]{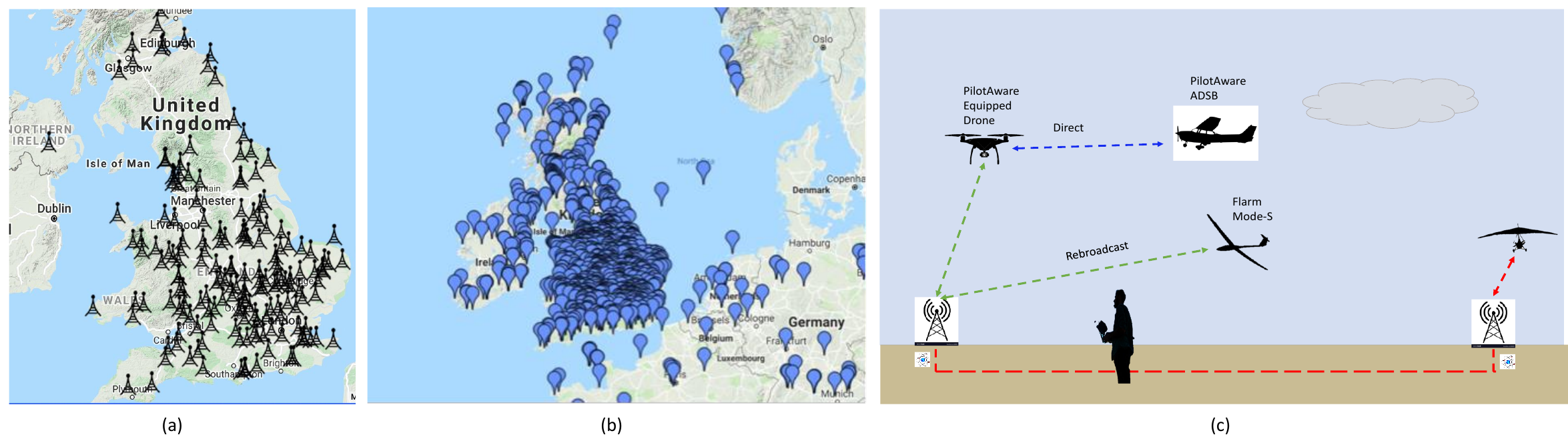}
	\end{center}
	\caption{PilotAware UK ATOM GRID (left); Radar 360 1090 ADS-B and MLAT coverage (middle); Data transmission (right).}
	\label{fig:PrestonMap1}
\end{figure*}

\begin{figure}[htp]
	\begin{center}
		\includegraphics[width=0.35\textwidth]{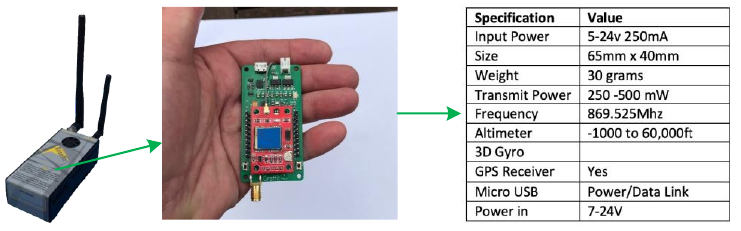}
	\end{center}
	\caption{PilotAware Rosetta-DX: detects all EC types with reduced size and power consumption for drone use.}
	\label{fig:Components}
\end{figure}

\subsection {Background}
\label{sec:Background}
The PilotAware ATOM-GRID Network (i.e., currently 250 UK ATOM stations) in the UK is depicted in Fig.~\ref{fig:PrestonMap1} a. It detects all aircraft transmitting an EC signal for specific types of aircraft. The stations, with a 50-60 km typical direct capture range are located at airfields, gliding sites, military sites, academic locations and private sites. The encrypted GRID network provides secure links to all stations. All stations are interconnected to share and re-transmit data and the distributed architecture provides greater redundancy than traditional technologies. In addition to the PilotAware GRID stations, data on the aviation frequency of 190 MHz from over 1300 stations as shown in Fig.~\ref{fig:PrestonMap1} b is accessed via the PilotAware system. This decentralised system using both centralised and decentralised instant real-time flight information is incorporated into the developed CM methodology in this study. The transmission of the data within this highly distributed network structure is illustrated in Fig.~\ref{fig:PrestonMap1} c. There is no standard EC device in use today. The miniaturised and lightweight version of the device, particularly, developed for UAVs in this research, is presented in Fig.~\ref{fig:Components}. Readers are referred to a YouTube video~\footnote{https://www.youtube.com/watch?v=JCpXdSFtHmU} for further information about how the PilotAware ATOM-GRID Network is operating. 


\subsection {Modules \& techniques in these modules}
\label{sec:Techniques}
\begin{figure*}[htp]
	\begin{center}
		\includegraphics[width=\textwidth]{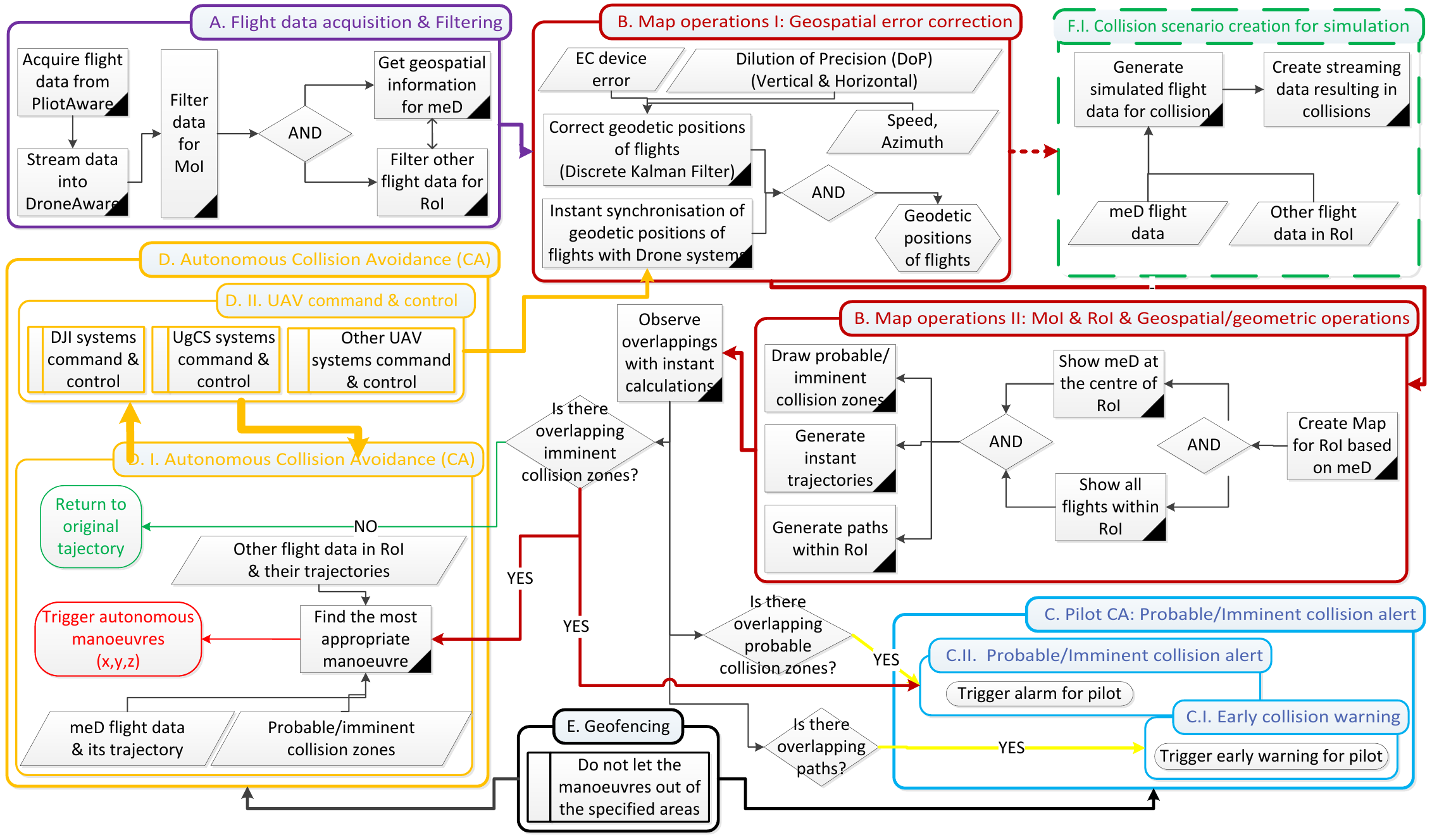}
	\end{center}
	\caption{Overall methodology with the modules presented in Fig,~\ref{fig:modules} and the developed techniques within these modules.}
	\label{fig:methodology}
\end{figure*}

UAVs are mainly categorised as fixed-wing and rotatory-wing. They have diverse characteristics and they are designed with particular features and onboard equipment concerning the missions they are expected to perform~\cite{9314128}. Compared with rotatory-wing UAVs, fixed-wing UAVs can carry more payload which allows carrying more fuel and can create space for a larger battery capacity. They also require less energy consumption during navigation because the wings can create a natural aerodynamic lift due to the air passing underneath with less thrust from the motors; therefore, fixed-wing UAVs generally have a longer range and can complete missions requiring longer flight durations~\cite{9314128}. However, fixed-wing design requiring large areas for take-off and landing along with steady forward movement is not suitable for urban use for many types of operations, which makes rotary-wing more appropriate for urban use with a cost of more energy consumption because of the excessive thrust from the motors to stay in the air~\cite{9314128}. Broadly speaking, fixed-wing UAVs have limitations with low manoeuvrability whereas they have longer flight time, which requires smoother CA manoeuvres with less diversion from the most recent trajectory. On the other hand, multi-copter UAVs have the advantage of very high manoeuvrability whereas they have limited flight time, which demands highly efficient CA manoeuvres involving the actions required for returning to the original trajectory. The CA techniques are mainly developed by looking out for these features along with the characteristics of the missions. The modules and the techniques (Fig.~\ref{fig:methodology}) within those modules are explored in the following subsections.

\begin{table*}[]
	\centering
	\caption{Data Streaming from the PilotAware system: Instant flight information. The value of the instant data capture in time, epoch (i.e., Unix timestamp), 1625226964, equals ``Fri Jul 02 2021 12:56:04 GMT+0100 (British Summer Time''. 
	}
	\label{tab:DataStreaming}
	\scalebox{0.825}{
		\begin{tabular}{|l|l|l|l|l|l|l|l|l|l|}
			\hline
			\rowcolor[HTML]{EFEFEF} 
			& \textbf{\begin{tabular}[c]{@{}l@{}}Time (epoch)\end{tabular}} & \textbf{\begin{tabular}[c]{@{}l@{}}Transponder\end{tabular}} & \textbf{\begin{tabular}[c]{@{}l@{}}Receiver\end{tabular}} & \textbf{Latitude} & \textbf{Longitude} &\textbf{\begin{tabular}[c]{@{}l@{}}Altitude (Feet)\end{tabular}} & \textbf{\begin{tabular}[c]{@{}l@{}}Heading\end{tabular}} & \textbf{\begin{tabular}[c]{@{}l@{}}Speed(knots/h)\end{tabular}} & \textbf{\begin{tabular}[c]{@{}l@{}}Call sign\end{tabular}} \\ \hline
			\cellcolor[HTML]{EFEFEF}\textbf{Example 1}                                                       & 1625226964    & 40717A                                                             & ADB                                                              & 38.962361         & 1.590027           & 3150              & 242                                                                   & 202                                                                           & EXS19TW                                                       \\ \hline
			\cellcolor[HTML]{EFEFEF}\textbf{Example 2}                                                       & 1625226965    & 406DF4                                                             & ADB                                                              & 56.53697          & -6.26629           & 7000              & 126                                                                   & 150                                                                           & GHIAL                                                         \\ \hline
		\end{tabular}
	}
\end{table*}

\subsubsection {Flight data acquisition \& filtering module}
\label{sec:filtering}
DTs, i.e., the virtual cyber-world embedded in the physical world, help map the real-time dynamic features of physical entities to the virtual world in multidimensional space. In this treatise, it is highly critical to update the flights’ geospatial positions precisely as frequently as possible for autonomous mission execution of FA-UAVs to adapt to dynamic environments and to guide pilots of PC-UAVs appropriately. In this direction, this module consists of two submodules: ``flight data streaming'' for real-time low latency EC data feeds acquired from all aircraft and ``flight data filtering'' for the processing and use of the most relevant flight data respective to a specific region (Fig.~\ref{fig:methodology} A).

\begin{figure}[htp!]
	\begin{center}
		\includegraphics[width=0.42\textwidth]{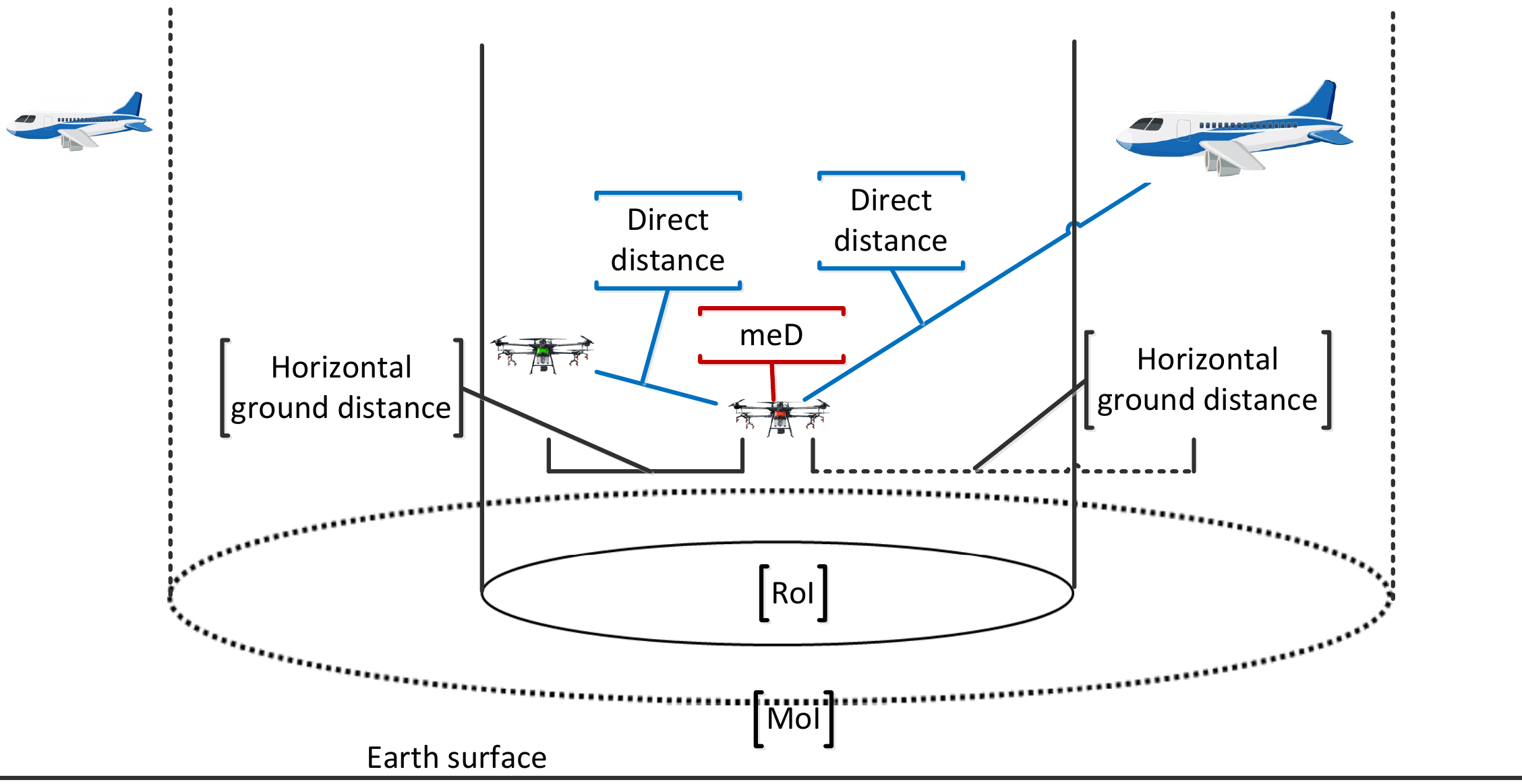}
	\end{center}
	\caption{Instant specification of $RoI$ and $MoI$ with respect to the changing localisation of $meD$ and filtering of flights accordingly.}
	\label{fig:RoI}
\end{figure}

\begin{algorithm}
	
	\caption{Threads for streaming flight data from the PliotAware system and filtering in DACM.}
	\label{alg:Streaming}
	\scriptsize
	\SetAlgoLined
	\KwData{\textbf{System input:} PilotAwareServerIP \& PortIP \& $meD$.ID}
	\KwData{\textbf{Instant input:} CurrentTime}
	\KwResult{$meD$.MoI \& $meD$.RoI \& $meD$.Data \& $MoI$.Flights.Data \& $RoI$.Flights.Data}
	
	$=>$ Creation of the UDP server\;
	UDPServer udpserver = new UDPServer()\;
	$=>$ Thread for streaming data from PilotAware system;\;
	Thread serverThread = new Thread(() $=>$ udpserver.Listen())\;
	$=>$ Thread for filtering\;
	Thread dataHandlerThread = new Thread(() $=>$ SubscribeToEvent(udpserver))\;
	\While{true}{
		$=>$ Start streaming\;
		[Flights.Data, $meD$.Data ]= serverThread.Start(PilotAwareServerIP, PortIP, $meD$.ID)\;
		$=>$ Start filtering\;
		[$meD$.MoI, $meD$.RoI, $MoI$.Flights.Data, $RoI$.Flights.Data] = dataHandlerThread.Start($meD$.Data, Flights.Data);
	}
	
	
\end{algorithm}

Two threads run in the background of the main application to feed the system with the flight data around the surrounding environment of the host drone, i.e., $meD$, as shown with the pseudo-codes in Alg.~\ref{alg:Streaming}. One of them manages the streaming of data from the PilotAware system into the DACM system using the format as exemplified in Table~\ref{tab:DataStreaming} via UDP connection whereas the other one filters the critical flights from this streaming and provides the main application with the crucial flight information within Map of Interest ($MoI$) and Region of Interest ($RoI$). More specifically, the streaming of data is filtered regarding $MoI$ (e.g., 20 miles away from $meD$ (i.e., the horizontal radius, $R$)) and restricted $RoI$ (i.e., ($R$ of $MoI$)/2) where $meD$ is placed at the centre of the geospatial map as displayed in Section~\ref{sec:Results}. The distances of other flights to $meD$ are calculated using latitude and longitude information concerning the horizontal ground distance as formulated in Table~\ref{eq-GrounDistance} to locate the flights either in $MoI$ or $RoI$ as delineated in Fig.~\ref{fig:RoI} where $MoI$ involves $RoI$ in the middle and $MoI$ = $4 x RoI$. In this way, not only is the number of flights related to $meD$ reduced, but also, the performance of the system is highly increased while carrying out the required geometric measurements, calculations and assessments e.g., path projection, collision assessment, determination of the collision-free aerospace and desired manoeuvres as elaborated in the following subsections. The filtering module helps the developed system run on ordinary computing devices efficiently and without demanding high computing resources. The delay in data transfer between the PilotAware and DACM system (i.e., $D_t = DA_t - PA_t$) is incorporated into the calculations to compensate for any delay in data acquisition where the first value (i.e., ICAO) in the streamed data (Table~\ref{tab:DataStreaming}) represents the time of the data acquired by the PlotAware system. The results acquired in our infrastructure using an orchestration of the UK's backhaul and fronthaul (i.e., crosshaul) communication technologies show that this time delay is around 1 sec using the distributed infrastructure (Section~\ref{sec:Background}) where the main servers of the PilotAware system are located in Holland.

\begin{table}[h!]
	\centering
	\caption{Calculation of the distances of other flights to $meD$ using latitude and longitude information concerning the horizontal ground distance.
	}
	\label{eq-GrounDistance}
	\scalebox{0.81}{
		\begin{tabular}{|l|} 
			\hline
			$GroundDistance = EarthRadius* 2 * \sin^{-1}( \sqrt{\sin^{2}(\frac{latDiff}{2})} +$ \\ $cos(meD_{radLat}) * \cos(other_{radLat})* \sqrt{\sin^{2}(\frac{lonDiff}{2})} );$\\
			\hline
			$DirectDistance = \sqrt{GroundDistance^2 + (meD_{alt} - other_{alt})^2}$;\\
			\hline
			where \hspace {5pt} $latDiff = meD_{radLat} - other_{radLat}$;\\ 
			$lonDiff = meD_{radLon} - other_{radLon}; \hspace {1pt} EarthRadius = 6378.137$;\\
			\hline
			\hline
		\end{tabular}
	}
\end{table}

\subsubsection {Map operations module}
\label{sec:mapOperations} 
The behaviouristic properties and spatial characteristics of flights need to be modelled correctly and visualised on a dynamic map for building a robust CM system. In this context, this module consists of four submodules: ”GPS location error correction”, “$MoI$ operations”, “$RoI$ operations” and “Geospatial and geometric operations” (Fig.~\ref{fig:methodology} B). While geometric calculations are performed, $f_{(x, y, z)}$ denotes the geospatial position of flights in the geographic global coordinate system, i.e., World Geodetic System (WGS-84), where $x$ and $y$ correspond to the horizontal axes of the flight in a map, i.e., longitude and latitude respectively whereas $z$ corresponds to the vertical axis, i.e., altitude. 

UAVs may encounter and cause danger if the signal in Global Positioning System (GPS) is weak or unavailable~\cite{ca92069d33d24c49b280c575f9b5c8ef}. Technically speaking, the GPS location error in the geodetic positions of flights due to dilution of precision (DoP), EC device error, ionospheric delays, and the number of available satellites and receiver clock errors resulting in tens of meters needs to be reduced to centimetre (cm) level for more accurate positioning. The enhanced positional SA paves the way for an effective CM system concerning the appropriate manoeuvres. Waypoints in flight routes result in a Gaussian distribution. Therefore, the pseudo-codes in Alg.~\ref{alg:KalmanFilter} within the ``GPS location error correction'' module update a flight state vector estimate based upon previous waypoints using Kalman Filter, in particular, discrete Kalman filter (DKF) in which linear stochastic mathematical calculations used in a way that minimises the mean of the squared error is incorporated into the system for modelling both autonomous and assisted navigation by estimating the future states. From a technical point of view, DKF estimates a process by using a form of feedback control~\cite{GWelch06}: i) the filter estimates the process state at some time and then obtains feedback in the form of (noisy) measurements, ii) as such, the equations for DKF fall into two groups: time update equations and measurement update equations where the time update equations are responsible for projecting forward (in time) current state and error covariance (Cov) estimates to obtain a priori estimates for the next time step whereas the measurement update equations are responsible for the feedback, i.e. for incorporating a new measurement into the priori estimate to obtain an improved posteriori estimate. The current estimate is recursively calculated using all of the	previous measurements involving the initial state using a priori and posteriori error covariance. In this way, GPS location information between the PilotAware and the DACM system, with the established DTs of aerial traffic, is synchronised on a cm basis using DKF by both smoothening the flight coordinates in an iterative reduced error in GPS precision and providing a timely prediction of the future flight state. This synchronisation based on not only the location, but also on the altitude is improved further by incorporating the local instant telemetry flight data (e.g., local GPS location, altitude, speed) of $meD$ using a telemetry listener thread developed within the methodology, which is mainly used while the CA commands are being implemented as explored in Sections~\ref{sec:pilotCollision} and~\ref{sec:autonomousCollision}. This thread feeds the system with local instant flight data and this data is compared with the data streamed from the PilotAware system by using the distance differences between the two systems for further synchronisation. To summarise, flight coordinates are smoothened by reducing the error in GPS precision and by providing a prediction of the future flight state using DKF and the telemetry data. From a mathematical point of view, DKF is initialised with Table~\ref{eq-Initialise}; State vector prediction and covariance are measured in Table~\ref{eq-State-vector-prediction-and-covariance}; Kalman gain factor (GF) and correction in the observation are obtained in Table~\ref{eq-Kalman-gain-factor}. 
\begin{algorithm}[t!]
	\scriptsize
	\caption{GPS error correction using DKF and synchronisation using the telemetry listener.} 
	\label{alg:KalmanFilter}
	\SetAlgoLined
	\KwData{\textbf{System input:} DoP, ECError}
	\KwData{\textbf{Instant input:} ($meD$.Data \& $RoI$.Flights.Data) $<==$ (Alg.~\ref{alg:Streaming}) \& meD.TelemetryData $<==$ (Telemetry.listener)}
	\KwResult{$meD$.CorrectedLocation \& $meD$.SyncronisedLocation \& $RoI$.Flights.CorrectedLocation \& $RoI$.Flights.SyncronisedLocation}
	
	\While{true}{
		$=>$ Correct the GPS error of $meD$\;
		[$meD$.CorrectedLocation] = KalmanFilter ($meD$.InitialState (x,y,z), $meD$.Speed, $meD$.Azimuth, DoP, ECError)\;
		$=>$ Correct the GPS errors of the flights within RoI\;
		[$RoI$.Flights.CorrectedLocation] = KalmanFilter ($RoI$.Flights.InitialState (x,y,z), $RoI$.Flights.Speed, $RoI$.Flights.Azimuth, DoP, ECError)\;
		$=>$ GPS location synchronisation between $meD$ and other flights\;
		[$meD$.SyncronisedLocation, $RoI$.Flights.SyncronisedLocation] = SyncroniseRoIFlights($meD$.TelemetryData.Location, $RoI$.Flights.CorrectedLocation)\;		
	}	
\end{algorithm}

\begin{table}[h!]
	\centering
	\caption{initialisation of DKF.
	}
	\label{eq-Initialise}
	\scalebox{0.81}{
		\begin{tabular}{|p{10cm}|} 
			\hline
			$SVE = observation_{mat}^{-1} * Flight(i)_{(x,y,z)};$\\
			\hline
			$hor_{error} = ECError * DoP_{hor};$\\
			\hline
			$ver_{error} = ECError * DoP_{ver};$\\
			\hline
			$Merror_{cov} = (hor_{error}, 0, 0; 0, hor_{error}, 0; 0, 0, ver_{error})$;\\
			\hline
			$Cov[SVE]= observation_{mat}^{-1} * Merror_{cov} * (observation_{mat}^{T})^{-1};$\\
			\hline
			where SVE is a priori state vector estimate before the addition of the new information and a posteriori vector estimate after the new information is measured and added; the default of $observation_{mat}$ is 3D square identity matrix; $Flight(i)_{(x,y,z)}$ is the observed geospatial vector of the Flight(i); $Merror_{cov}$ is geospatial measurement error; $Cov[SVE]$, 3D square matrix, is covariance of state vector estimate.\\
			\hline
			\hline
		\end{tabular}
	}
\end{table}

\begin{table}[h!]
\centering
\caption{State vector prediction and measurement of covariance ($Cov$).}
\label{eq-State-vector-prediction-and-covariance}
\scalebox{0.81}{
	\begin{tabular}{|p{10cm}|}
		\hline
		$SVE = ST_{mat}* SVE + input_{mat} * inputControl_{vec};$\\
		\hline
		$Cov[SVE] = ST_{mat}* Cov[SVE] * ST_{mat}^{T} + Perror_{mat}$;\\
		\hline
		where the default of $ST_{mat}$ (state transition matrix) is a 3D square identity matrix; the default of $input_{mat}$ (input matrix) is 3D square zero matrix; the default of $inputControl_{vec}$ is zero vector for Cartesian points; the default of $Perror_{mat}$ is 3D square zero matrix for processing noise covariance.\\
		\hline
		\hline
	\end{tabular}
}
\end{table}

\begin{table}[h!]
	\centering
	\caption{Measurement of Kalman gain factor (GF) and correction in the observation.}
	\label{eq-Kalman-gain-factor}
	\scalebox{0.81}{
		\begin{tabular}{|p{10cm}|}
			\hline
			$GF = Cov[SVE] * (observation_{mat})^{T} * (observation_{mat}*$ \\ 
			$Cov[SVE] *(observation_{mat})^{T} + Merror_{cov} )^{-1};$\\
			\hline
			$SVE = SVE + GF *(Flight(i)_{(x,y,z)} - observation_{mat} * SVE);$\\
			\hline
			$Cov[SVE] = Cov[SVE] - GF * observation_{mat} * Cov[SVE];$\\
			\hline
			where the default of $Merror_{mat}$ is a square zero matrix for measuring noise covariance and $SVE$ results in a new corrected point (x,y,z) (i.e., an improved estimate of $SVE$) based on observation along with its covariance, $Cov[SVE]$. The execution of the DKF within an example video is in the supplements.\\
			\hline
			\hline
		\end{tabular}
	}
\end{table}

Map operations are dynamically performed regarding two regions i) $MoI$ and ii) $RoI$ considering that these regions are highly dynamic regarding the high speed of the flights (Section~\ref{sec:filtering}), particularly aeroplanes. The $RoI$ operations are the backbone of the system where the flights in this region are in close proximity to $meD$ and the other flights out of $RoI$ but within $MoI$ are the potential flights approaching $RoI$. There are more uncertainties in the trajectories and geometry formations of drones compared to aeroplanes where they are deployed for many different types of versatile missions in which the routes along with trajectories are highly volatile with constantly changing projected/predicted path-loss, in particular, while rotatory-wing based UAVs are performing the missions with nonlinear trajectories. Sometimes their missions are dependent on the ground such as filming, landmine detection, search and rescue and surveillance; sometimes independent from the ground, i.e., air-dependent between take-off and destination such as package delivery. Hence, particular approaches specific to the characteristics of the drones and their missions shall be developed to determine the conflict conditions between vehicles and deconflicting manoeuvres to avoid potential collisions. The main goal of CM is to keep flights away from hazardous regions starting from the very early stages, which leads to minimising the future collision risk significantly but keeping the efficient use of drones in mind due to battery constraints. In this sense, first, predicted paths with the projected trajectories concerning the velocity vector of the flights are generated for conflict management to perform deconflicting where the paths are overlapping one another. In this way, the current path for $meD$ is modified to separate the overlapping paths of flights leading to free-of-conflict routes, i.e., Flight of Safety Route ($FoSR$). Second, SA through horizontal and vertical planes is provided with active flights around $meD$ and collision risk assessment is processed relying on their trajectory predictions. The three risk zones for determining the seriousness of the collision risk are i) Probable-collision-Risk zone ($PcRz$) as the low-collision-risk-aerospace, ii) Imminent-collision-Risk zone ($IcRz$) as the high-collision-risk-aerospace where $PcRz > IcRz$ and iii) collision-Free zone ($cFz$) as clear of $PcRz$ and $IcRz$. These zones are measured and drawn on the map for every flight involving manned aeroplanes within $RoI$. These concepts are elaborated as follows.

\begin{table}[h!]
	\centering
	\caption{Formulas for the conflict geometry of $PCTD$ and $PcRz$ based on combined speeds of two aircraft (Fig.~\ref{fig:PCTZones})}
	\label{eq1}
	\scalebox{0.81}{ 
		\begin{tabular}{|p{10cm}|}
			\hline
			$PcRz_{f1} ==> (Dist_{f1} \leq PCTD_{meD} + PCTD_{f1}) \hspace {3pt} \& \hspace {3pt}$\\
			$(Dist_{f1} > ICTD_{meD} + ICTD_{f1});$\\
			\hline
			where \hspace {1pt} $(0 < P_c \leq 0.5) \hspace {3pt} \& \hspace {3pt} (PCTD = 2 \hspace {3pt}X\hspace {3pt} ICTD)$\\
			\hline
			\hline
		\end{tabular}
	}
	\begin{center}
		\includegraphics[width=0.4\textwidth]{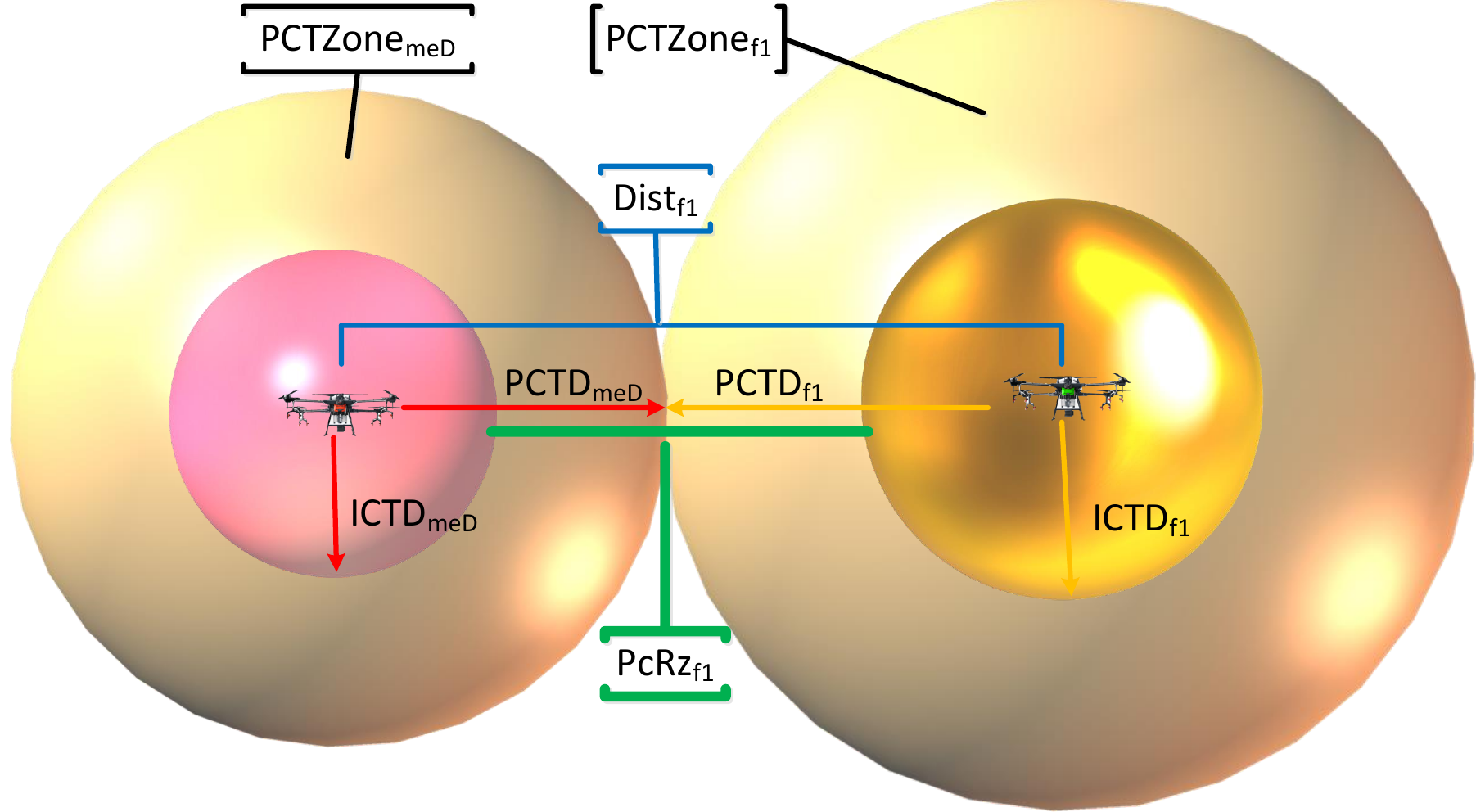}
	\end{center}
	\captionof{figure}{Conflict geometry of $PCTD$ and $PcRz$: The yellow zone for each flight represents $PCTZone$. The red zone for $meD$ and orange zones for other flights represent $ICTZones$ (Table~\ref{eq1}).}
	\label{fig:PCTZones}
\end{table}

\begin{table}[h!]
	\centering
	\caption{Measurement for the Conflict geometry of $ICTD$ and $IcRz$ based on combined speeds of two aircraft (Fig.~\ref{fig:ICTZones}).}
	\label{eq23}
	\scalebox{0.81}{
		\begin{tabular}{|p{10cm}|}
			\hline
			$IcRz_{f1} ==>	(Dist_{f1} \leq ICTD_{meD} + ICTD_{f1}) \hspace {3pt} \& \hspace {3pt} (Dist_{f1} > 0);$\\
			\hline
			where \hspace {10pt} $(0.5 < P_c \leq 1) \hspace {3pt} \& \hspace {3pt} (PCTD = 2 \hspace {3pt}X\hspace {3pt} ICTD);$\\
			\hline
			$ICTZone = \frac{4}{3}\pi (ICTD)^3;$\\
			\hline
			$PCTZone = (\frac{4}{3}\pi (PCTD)^3) - ICTZone = \frac{28}{3}\pi (ICTD)^3;$\\	
			\hline
			\hline
		\end{tabular}
	}
	\begin{center}
		\includegraphics[width=0.3\textwidth]{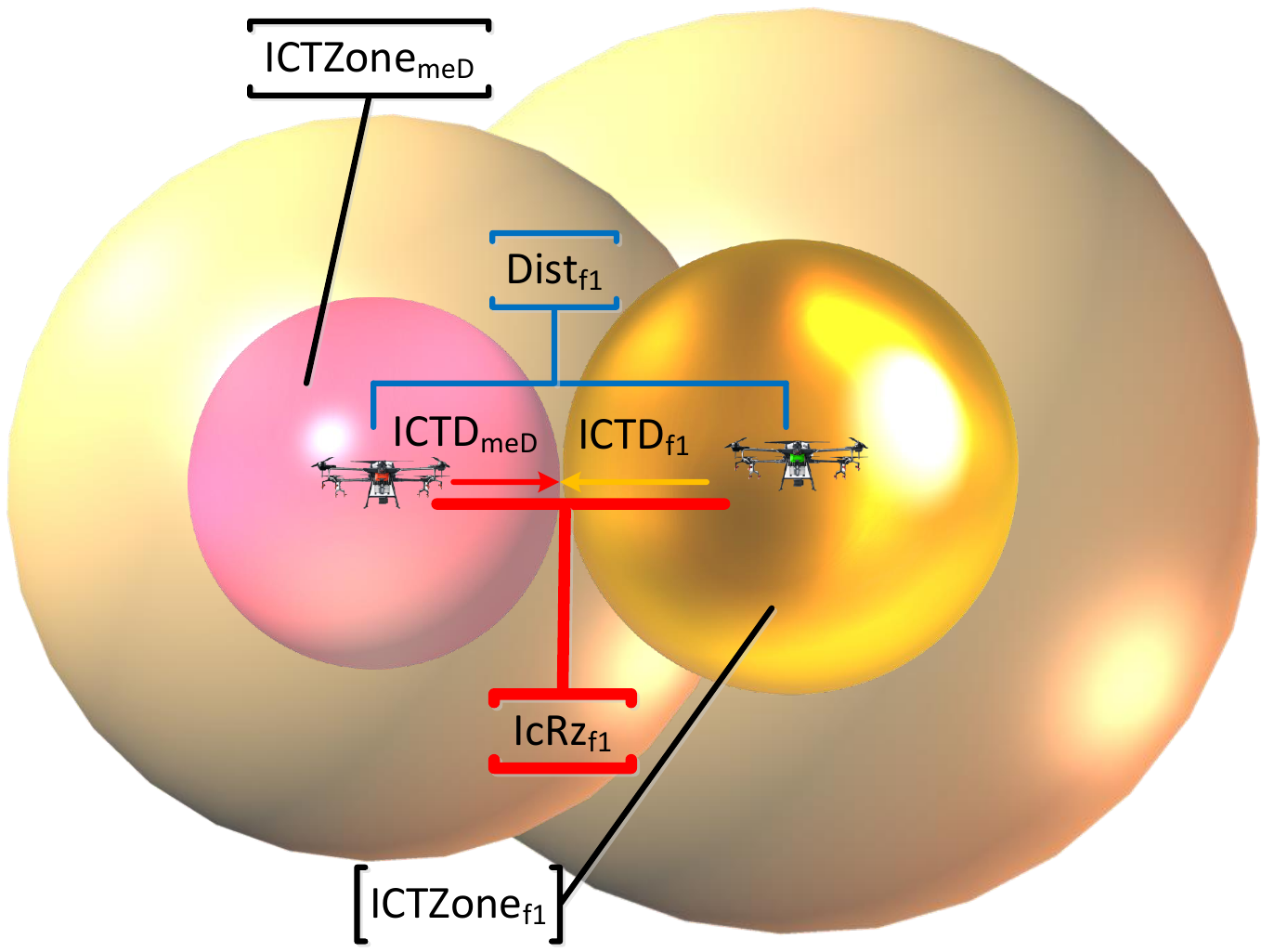}
	\end{center}
	\captionof{figure}{Conflict geometry of $ICTD$ and $IcRz$ (Table~\ref{eq23}).}
	\label{fig:ICTZones}
\end{table}

\begin{algorithm}
	\caption{Modelling of probable collision risk zone ($PcRz$) \& Imminent collision risk zone ($IcRz$).}
	\label{alg:PcRzIcRz} 	
	\scriptsize
	\SetAlgoLined
	\KwData{\textbf{System input:} $PCTD$ \& $ICTD$}
	\KwData{\textbf{Instant input:} $meD$.Data \& $RoI$.Flights.Data \& $RoI$.UpdatedMap \& ($meD$.SyncronisedLocation \& ($meD$.CorrectedLocation \& $RoI$.Flights.CorrectedLocation \& $RoI$.Flights.SyncronisedLocation $<==$ (Alg.~\ref{alg:KalmanFilter})}
	\KwResult{$RoI$.Flights.PcRz \& $RoI$.Flights.IcRz}
	\While{true}{ 
		$=>$ Draw geometric calculations of $meD$ on the map\;
		[$meD$.PCTZone, $meD$.ICTZone] = calculateDrawMap-CTZones($meD$.CorrectedLocation, $PCTD$, $ICTD$)\;
		$=>$ Show other flights on the map and draw geometric calculations\;
		$=>$ Detect $PcRz$ and $IcRz$\;
		\ForEach{($RoI$.Flights)}{
			$=>$ Calculate instant direct distance (Table~\ref{eq-GrounDistance}) between two flights\;
			[DirectDistance\textsubscript{$RoI$.Flights(i)}] = calculateDirectDist ($meD$\textsubscript{(X,Y,Z)}, $RoI$.Flights(i)\textsubscript{(X,Y,Z)})\;
			$=>$ Draw geometric calculations on the map\;
			[$RoI$.Flights(i).PCTZone, $RoI$.Flights(i).ICTZone] = calculateDrawMap-CTZones($RoI$.Flights(i).CorrectedLocation, $PCTD$, $ICTD$)\;			
			$=>$ Determine if there is $PcRz$\;
			[isThere-PcRz, $RoI$.Flights(i).$P_c$, $RoI$.Flights(i).PcRz] = determine-PcRz($meD$.PCTZone, $RoI$.Flights(i).PCTZone, DirectDistance\textsubscript{$RoI$.Flights(i)})\;
			$=>$ Determine if there is $IcRz$\;
			[isThere-IcRz, $P_c$, $RoI$.Flights(i).IcRz] = determine-IcRz($meD$.ICTZone, $RoI$.Flights(i).ICTZone, DirectDistance\textsubscript{$RoI$.Flights(i)}\;

			\uIf{(isThere-PcRz == false) \&\& (isThere-IcRz == false)}{
				$=>$ No collision risk\;
			}
			\uElseIf{(isThere-PcRz == true)}{ $=>$ Probable collision risk\;
				\uIf{PC-UAV}{
					warnPilot($P_c$, PCTZone)\;
				}
				\Else{ $=>$ Warn base station\;
					warnBS($P_c$, PCTZone)\;
				}								
			}
			\Else{ $=>$ Imminent collision risk\;				
				Alg.~\ref{alg:CollisionAvoidance} $<==$ ($meD$.Data, $RoI$.Flights(i).Data, $meD$.SyncronisedLocation, $RoI$.Flights(i).SyncronisedLocation, $RoI$.Flights(i).IcRz, $RoI$.Flights(i).PcRz, $RoI$.Flights(i).IcRz, $RoI$.Flights(i).$P_c$)\;																			 
			}
		}
		
	}
\end{algorithm}

\textbf{a) Probable collision travel distance ($\textbf{PCTD}$) \& Imminent collision travel distance ($\textbf{ICTD}$) \& $\textbf{PcRz}$ \& $\textbf{IcRz}$:}
The instant modelling of collision assessment zones is difficult for nonlinear trajectories and variable speeds of drones as explained earlier.
To mitigate these concerns, various geometric techniques were developed depending on the constraints and requirements. These techniques were used to estimate $PcRz$ and $IcRz$ leading to determining the probability of collision ($P_c$) corresponding to the proximity of the flights -- time to collision (TTC). The pseudo-codes for modelling those collision risk zones are presented in Alg.~\ref{alg:PcRzIcRz}.

Depending upon the size, capabilities and characteristics of the aircraft an appropriate policy has to be developed whereby hypothetical repulsive force fields are dynamically instantiated around UAVs with suitable magnitudes ensuring correct separation distances are maintained. $PCTD$ and $ICTD$ are defined by the user regarding time-based distance (e.g., TTC) in seconds as travel distance ($TD$) concerning direct horizontal and vertical distances as illustrated in Figs.~\ref{fig:PCTZones} and~\ref{fig:ICTZones}. Multiple collision risk assessment zones as many as the number of the flights within $RoI$ are generated regarding their location, direction and $TD$ in metres (m), in a broader perspective, using the geospatial position and angular velocity vectors of the flights. The yellow spherical zone around each flight represents the probable collision travel zone ($PCTZone$) whereas the red zone around $meD$ and orange zones around other flights represent imminent collision travel zones ($ICTZones$). These instant highly varying zones are calculated and drawn on the map readily as the flight data streams as explained in Section~\ref{sec:filtering}. The higher the speed, the larger the zones with a dynamic collision radius, $R$, which leads to the same predetermined TTC proximity zone concerning the navigation time despite varying distances being obtained. For instance, the zones for $f_1$ are larger than the zones of $meD$ where the speed of $f_1$ is twice the speed of $meD$. A collision probability value, $P_c$, indicates the seriousness of the risk. The closer the flights the higher the probability of collision. The overlapping $PCTZones$ generate $PcRz$ as formulated in Table~\ref{eq1} (Fig.~\ref{fig:PCTZones}) where $P_c$ is between 0 and 0.5. The overlapping $ICTZones$ generate $IcRz$ as formulated in Table~\ref{eq23} (Fig.~\ref{fig:ICTZones}) where $P_c$ is between 0.5 and 1 with highly increasing collision risk. From the geometric standpoint, $PcRz$ occurs where the yellow spherical aerial volumes (i.e., $PCTZones$) of two flights start to invade each other's district whereas $IcRz$ occurs where their red/orange spherical aerial volumes (i.e., $ICTZones$) intertwine with each other. The volume calculation of these two zones is formulated in Table~\ref{eq23}.

\textbf{b) Collision-Free zone ($\textbf{cFz}$) \& collision-Risk-Less zone ($\textbf{cRLz}$):} 
\label{sec:cFz}
Real-time modelling of $cFz$ and $cRLz$ within $RoI$ is conducted for determining the appropriate autonomous manoeuvres regarding the current $IcRz$. These safe regions to manoeuvre are identified in line with the conflict regions with other flights to avoid any further $IcRz$, where multiple conflicts in trajectories between multiple flights are possible and collision risk assessment is processed with each flight. In this regard, the aerospace not covered by both $PCTZones$ and $ICTZones$ are considered as $cFz$ whereas the aerospace not covered by $ICTZones$, but by $PCTZones$ are considered as $cRLz$. Autonomous manoeuvres to avoid collisions are executed considering these designated aerial volumes with prioritisation of using $cFz$ as elaborated in Sections~\ref{sec:pilotCollision} and~\ref{sec:autonomousCollision}.

\begin{figure}[tp!]
	\begin{center}
		\includegraphics[width=0.35\textwidth]{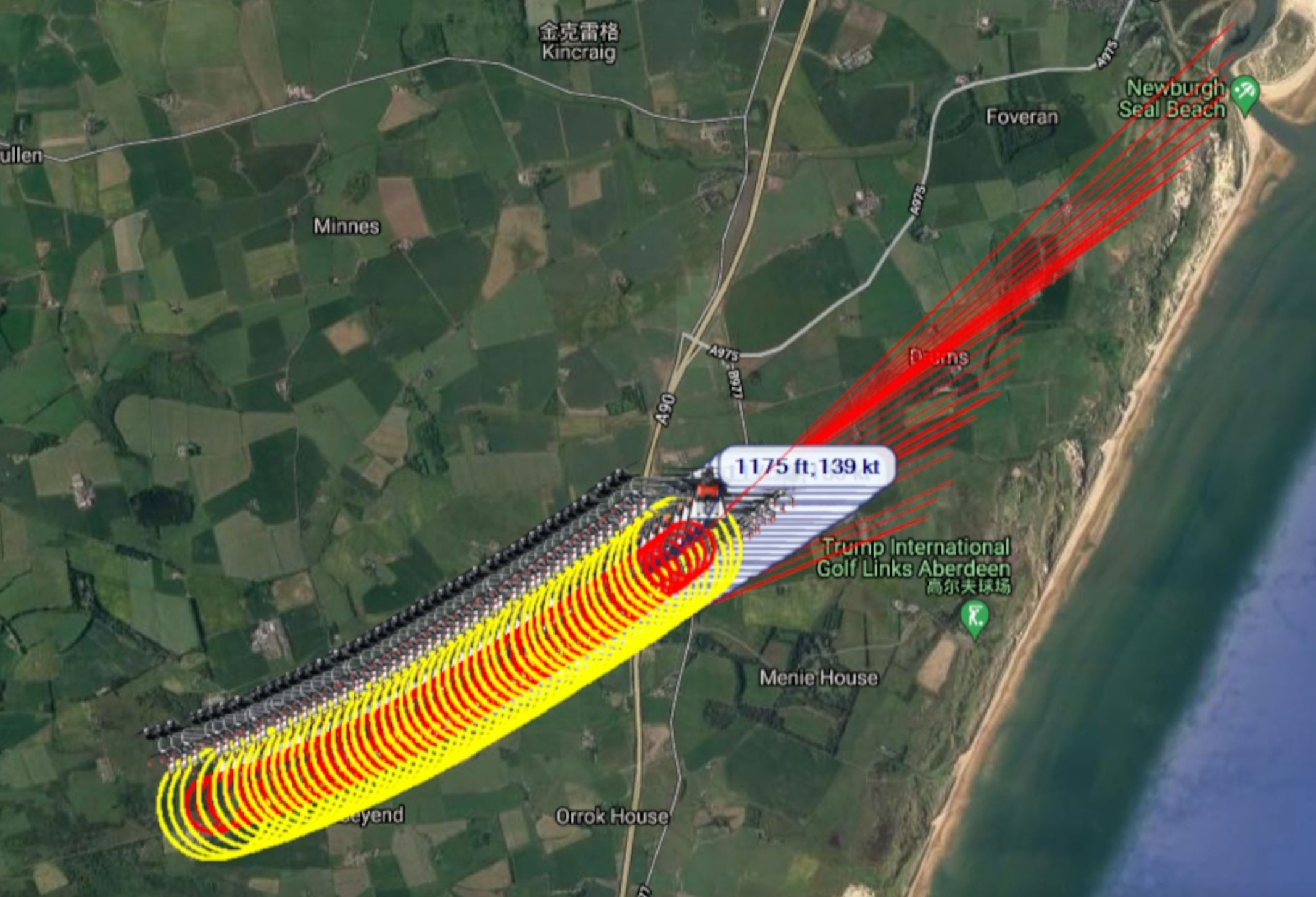}
	\end{center}
	\caption{Generation of future waypoints: safety travel tube.}
	\label{fig:waypointsFuture}
\end{figure}

\textbf{c) Flight of Safety Route ($\textbf{FoSR}$):}
\label{sec:FoSR}
The direction per instant trajectory is drawn with a red line for $meD$ and orange lines (Fig.~\ref{fig:realTest}) for the other flights. Path prediction concerning the instant occurring trajectories is incorporated into the system for pilot early collision risk assessment (Section~\ref{sec:pilotCollision}). The consecutive travel distances between the waypoints with the flight coordinates, $f_{(x, y, z)}$, along with the speed acquired from the PilotAware system are processed to predict the future waypoints of the flights within $RoI$. In other words, the routes are predicted using the patterns of the consecutive occurring waypoints (i.e., the instant linear or nonlinear trajectories) whose projected conjunctions are expected to result in a steady linear or nonlinear path, which is exemplified in Section~\ref{sec:pilotCollision} (Fig.~\ref{fig:waypointsFuture}) with a video. The flights out of $RoI$ are excluded from the calculations if their trajectory directions with orange lines are not heading into $RoI$ to increase the efficacy of the system with reduced processing. The predicted paths are recalculated as the flight data streams into the system, but are not drawn on the map in order not to make the screen complicated with many drawings. $FoSR$ for $meD$ is considered as the predicted path with no overlapping conflicting predicted routes of other flights with no $PcRz$ and $IcRz$.

\begin{algorithm}
	\caption{CM policy with no collision risks with other flights by sweeping for $cFz$ or $cRLz$.}	
	\label{alg:CollisionAvoidance}
	\scriptsize
	\SetAlgoLined
	\KwData{\textbf{System input:} $meD$.Property}
	\KwData{\textbf{Instant input:} $meD$.Data, $RoI$.Flights(i).Data, $meD$.SyncronisedLocation, $RoI$.Flights(i).SyncronisedLocation, $RoI$.Flights(i).IcRz, $RoI$.Flights(i).PcRz, $RoI$.Flights(i).IcRz, $RoI$.Flights(i).PcRz, $RoI$.Flights(i).$P_c$}
	\KwResult{$meD$.collisionFreeTrajectory}
	
	$=>$ Find the nearest CA waypoint within $cFz$ or $cRLz$ by sweeping\;
	\While{isThere-IcRz}{
		$=>$ Determine the new waypoints with no new collision risks\;
		[$meD$.newVelocityVector (newHeading, newSpeed, targetCoordinates (x,y,newAltitude))] = MenouevreCA ($meD$.Property(model), $meD$.Property(OriginalTrajectory), $meD$.Data, $RoI$.Flights(i).Data, $RoI$.Flights(i).ICTZone) (Alg.~\ref{alg:CA-AI})\;	
		
		[$meD$.collisionFreeTrajectory ($meD$.NewWaypoints)] = addVelocityVector($meD$.newVelocityVector)\;	
	}
	$=>$ Return back to the preplanned route\;
	\While{isThere-PcRz}{
		wait; $=>$ Wait for $cFz$ to return back to trajectory\;
	}
	\uIf{($meD$.Property(ControlType) == "Autonomous") \&\& ($meD$.Property(MissionType) == "GroundBased")}{
		$=>$ Return to the trajectory where CA starts\;
		returnOriginalRoute-GB ($meD$.Property(model), $meD$.Property(OriginalRoute), $meD$.StartLocationCM)\;
	}
	\uElseIf{($meD$.Property(ControlType) == "Autonomous") \&\& $meD$.MissionType == "AirBased"}{
		$=>$ Return to the nearest point in the preplanned route\;
		returnOriginalRoute-AB ($meD$.Property(model), $meD$.Property(OriginalRoute), $meD$.CurrentLocation)\;							
	}
	\Else{ $=>$ ($meD$.Property(ControlType) == "Pilot")\;	
		$=>$ Wait for the pilot to manoeuvre\;			
		changeMode($meD$.Property(mode) = "Joystick")\;
	}
\end{algorithm}

\subsubsection {Pilot CA module}
\label{sec:pilotCollision}
The main goal of CM is to keep flights away from hazardous regions with conflict-free solutions at the very early stage by keeping each flight well clear of other traffic considering their current trajectories and preplanned routes. Currently, UAVs do not have a system that will warn an operator of an impending collision with other airborne vehicles~\cite{5937290} using broader local coverage. In this module (Fig.~\ref{fig:methodology} C), SA through horizontal and vertical planes is provided with active flights around $meD$ and early collision risk assessment is processed considering their trajectory predictions in relation to that of $meD$. From a linear formulation technical viewpoint, the flights' time-invariant connectivity of previous waypoints, their trajectory headings and their $PCTZones$ and $ICTZones$ are used to generate the projections of future linear or nonlinear waypoints (e.g., kinematic cartesian virtual points (CVPs)) leading to a safety travel tube (STT) as depicted in Fig.~\ref{fig:waypointsFuture}. It is noteworthy to emphasise that "nonlinear waypoints" are the product of nonlinear trajectories as elaborated earlier in Section~\ref{sec:mapOperations}. Readers are referred to the video, TravelTube, in the supplements for the implementation based on the most recent consecutive waypoints. $meD$ is said to be within $FoSR$ where there are no overlapping STTs. 
The DACM system is responsible for instructing the flight controller to maintain UAV separation from conflicting regions with a conflict resolution mechanism and to avoid imminent collisions autonomously. The pilot of $meD$ is first informed about the approaching collision risk if its STT are intersecting at a point of the STT segment of any other flight considering the conflict geometry. The system, second, warns for the probable collision risks (within $PcRz$) as illustrated in Fig.~\ref{fig:PCTZones}. Third, it alerts for imminent collision risks (within $IcRz$) as illustrated in Fig.~\ref{fig:ICTZones} along with autonomous deconflicting manoeuvres by providing the pilot with an extra layer of security. 

The pseudo codes of the CM policy and the concept of autonomous CA are provided in Algs.~\ref{alg:CollisionAvoidance} and~\ref{alg:CA-AI} respectively. CA for PC-UAVs mainly relies on the human operator's remote commands regarding the early and probable collision risks. Pilots are expected to take proper action through other possible routes when they are informed about the predicted conflicting routes and current trajectories along with the heading, altitude and speed of all flights (Fig.~\ref{fig:selfMan}) are displayed on the map on a real-time basis with real-time tracking of flights as explained in Section~\ref{sec:filtering}. The most available safety corridor route for avoiding $PcRz$ and $IcRz$, i.e., $FoSR$s, can be clearly distinguished by the pilot through $cFz$ \& $cRLz$ regarding vertical and horizontal planes. The manoeuvrability performance of the pilot is taken into consideration in specifying $PCTD$ that allows effective manoeuvring (e.g., 12 sec $TD$). No autonomous action is advised where the collision risks are highly low regarding the large distances for the future conflicting flight paths regarding STTs and probable collision risks. However, appropriate autonomous manoeuvres are executed i) for imminent collision risks within an unsafe response range (i.e., within $IcRz$) as elaborated in Section~\ref{sec:autonomousCollision} for FA-UAVs if the pilot has not taken any action during the prior aforementioned probable collision risk warnings and ii) to correct the pilot's operational errors leading to instant imminent collision risks. 

\begin{algorithm} [t!]
	\caption{MenouevreCA: CA using a bespoke pairwise collision risk technique.}
	\label{alg:CA-AI}
	\scriptsize
	\SetAlgoLined
	\KwData{\textbf{System input:} $meD$.Property \& $meD$.OriginalRoute \& $meD$.MissionType \& $meD$.ControlType \& $meD$.Property}
	\KwData{\textbf{Instant input:} $meD$.Property(model), $meD$.Property(OriginalTrajectory), $meD$.Data, $RoI$.Flights(i).Data, $RoI$.Flights(i).ICTZone, DirectDistance\textsubscript{$RoI$.Flights(i)}}
	\KwResult{$meD$.newVelocityVector (newHeading, newSpeed, targetCoordinates (x,y,newAltitude))}
	$=>$ Determine the new heading for for CA manoeuvre \;
	diffHeading = $RoI$.Flights(i).Heading - $meD$.Heading \;
	\uIf{(diffHeading $<$ 0)}{ 		
		HeadingCA = $meD$.Heading - $meD$.Property(DivertionAngle)\;
		\uIf {HeadingCA $<$ 0}{
			HeadingCA = HeadingCA + 360\;
		}		
	}
	\Else {
		HeadingCA = $meD$.Heading + $meD$.Property(DivertionAngle)\;
		\uIf {(HeadingCA $>$ 360)}{
			HeadingCA = HeadingCA - 360 \;
		}
	}	
	
	$IcRz$\textsubscript{$RoI$.Flights(i)} = $meD$.ICTD + $RoI$.Flights(i).ICTD\;
	
	$=>$ Determine the desired manoeuvre distance for CA manoeuvre\;
	$meD$.MoveDistance = $IcRz$\textsubscript{$RoI$.Flights(i)} - DirectDistance\textsubscript{$RoI$.Flights(i)}\;
	
	$=>$ Determine the vertical move ability in metres regarding the ability of the drone and the $meD$.MoveDistance\;
	[upwardChange, downwardChange, newSpeed] = CalculateVerticalChangeSpeed ($meD$.MoveDistance, $meD$.property(UpwardMove, DownwardMove))\; 
	
	$=>$ Determine upward or downward vertical direction\; 	
	\uIf{($meD$.Altitude $<$ $RoI$.Flights(i).Altitude)}{ 
		VerticalMv = - downwardChange\; 		
	}
	\Else {
		VerticalMv = upwardChange\; 
	} 
	
	$meD$.NewAltitude = $meD$.Altitude + VerticalMv\;
	
	$=>$ Determine the new CA coordinates\;
	[$meD$\textsubscript{(x1,y1,z1)} ] = GetPointByDistanceAndHeading($meD$\textsubscript{(x,y,z)}, $meD$.HeadingCA, $meD$.MoveDistance, , $meD$.NewAltitude)\;
	return $meD$.newVelocityVector (HeadingCA, newSpeed, targetCoordinates (x1,y1,z1))\;		
\end{algorithm}

\subsubsection {Autonomous CA module}
\label{sec:autonomousCollision}
Fully self-operating systems are human-out-of-the-loop systems that single-handedly determine the right course of action when given an autonomous task~\cite{9637501}. In an autonomous mission, a self-operating UAV is fully aware of its mission starting point, altitude, current location, navigation speed, middle waypoints, heading and target location. Reliable operations of UAVs under imminent collision risks are essential. Safety has to be ensured in autonomous navigation by dealing with sudden changes in the dynamic UAV environment concerning the aforementioned characteristics of UAVs. Safe zones such as $cFz$ and $cRLz$ (Section~\ref{sec:mapOperations} C) are used to enable safe autonomous operations/manoeuvres for avoiding the imminent mid-air collision risks without causing further risks for both $meD$ and other nearby flights, including manned aeroplanes. An effective CM methodology with energy-efficient -- i.e., the minimal spatial deviation from the preplanned trajectory -- reactive 3D CA manoeuvres is implemented in this study by considering both quick adaptations to new dynamic environments during manoeuvres and turning back to the original trajectory as smoothly as possible. 
The conceptual pseudo-codes of this module (Fig.~\ref{fig:methodology} D) which aims to react predictably to the dynamic environment are given in Algs.~\ref{alg:CollisionAvoidance} and~\ref{alg:CA-AI}. The system triggers an imminent collision risk if their $ICTZones$ satisfy $IcRz$ condition (Fig.~\ref{fig:ICTZones}). $meD$ needs to change its trajectory into a safe zone when there is an imminent collision risk regarding the proximity to collision (i.e., $P_c$). $meD$ is autonomously directed by the system with appropriate autonomous avoidance manoeuvres such as 3D direction deviation, acceleration/deceleration and/or change of the altitude via the drone's telemetry using the drone control systems (e.g., UgCS, DJI). The main objective of the manoeuvring algorithm is to take the $meD$ out of $ICTZones$ as quickly as possible, which depends on the capabilities of the $meD$, which requires the acceleration. Safe regions to manoeuvre are determined concerning the conflict regions with the other close-range flights not to cause further risks where multiple conflicts between multiple flights during changing trajectories are possible and collision risk assessment is performed with each flight in $RoI$ instantaneously. To be concise, ``RoI.Flights(i).Data'' in Alg.~\ref{alg:CollisionAvoidance} indicates any flight that creates a risk of collision in its environment and the input into~Alg.~\ref{alg:CA-AI} is any flight (i) that is causing collision risks. In other words, in the first instance, the CA manoeuvre is implemented through $cFz$ or $cRLz$ to avoid further collision risks and Alg.~\ref{alg:CA-AI} is triggered for multiple flights in ROI if their $ICTZones$ intersect with meD considering SA within ROI. 

\begin{table}[h!]
	\centering
	\caption{Formulas for CA manoeuvres (Fig.~\ref{fig:AICA} and Alg.~\ref{alg:CA-AI})}
	\label{eqMan}
	\scalebox{0.76}{
		\begin{tabular}{|p{10.7cm}|}
			\hline
			$\textbf{CALatR} = \sin^{-1}(\sin(latR)*\cos(distCARadius) + $\\ 
			$\cos(latR)*\sin(distCARadius) * \cos(bearingCAR));$\\
			\hline
			$\textbf{CALonR} = LonR+\tan^{-1} ((\sin(bearingCAR)*\sin(distCARadius) *$\\
			$\cos(latR))/ (\cos(distCARadius) - \sin(latR) * \sin(meCALat));$\\
			\hline
			$\underline{\textbf{meDCALat}} = CALatR * (180/\pi);\underline{\textbf{meDCALon}} = CALonR * (180/\pi);$\\
			\hline
			$\underline{\textbf{meDCALat}} = CALatR * (180/\pi);\underline{\textbf{meDCALon}} = CALonR * (180/\pi);$\\
			\hline
			where \hspace {1pt} $latR = meLat * (\pi/180); lonR = meLon* (\pi/180);$\\
			$bearingCAR = newHeading*(\pi/180);$	\\
			$distCARadius = MoveDistance / EarthRadius$;\\
			\hline
			\hline
		\end{tabular}
	}
	\begin{center}
		\includegraphics[width=0.27\textwidth]{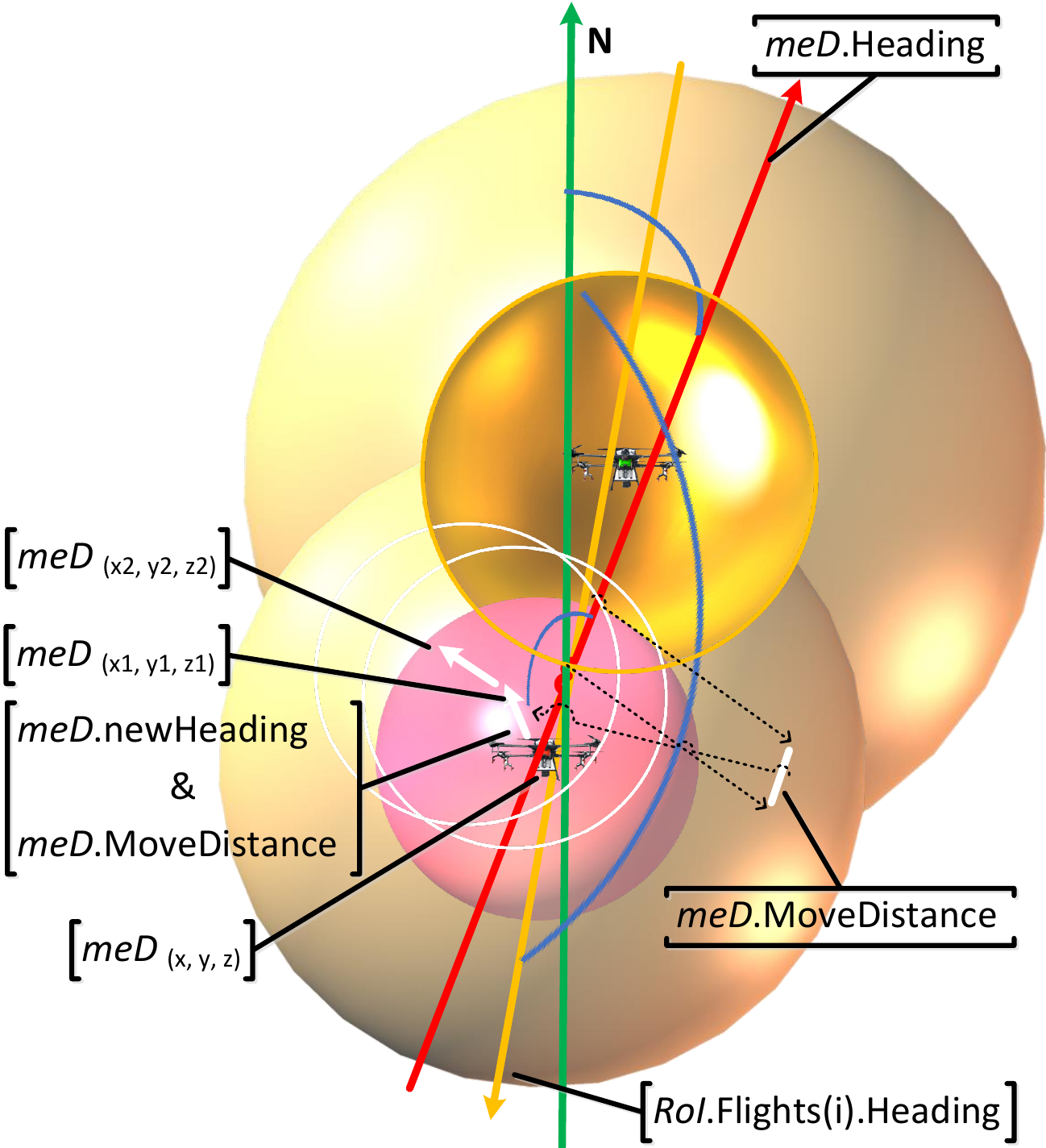}
	\end{center}
	\captionof{figure}{Geometric illustration of CA manoeuvres with Alg.~\ref{alg:CA-AI}. $meD.MoveDistance$ is corresponding to the invasion length of $ICTZones$ (Table~\ref{eqMan}).}
	\label{fig:AICA}
\end{table}

\begin{figure}[tp!]
	\begin{center}
		\includegraphics[width=0.35\textwidth]{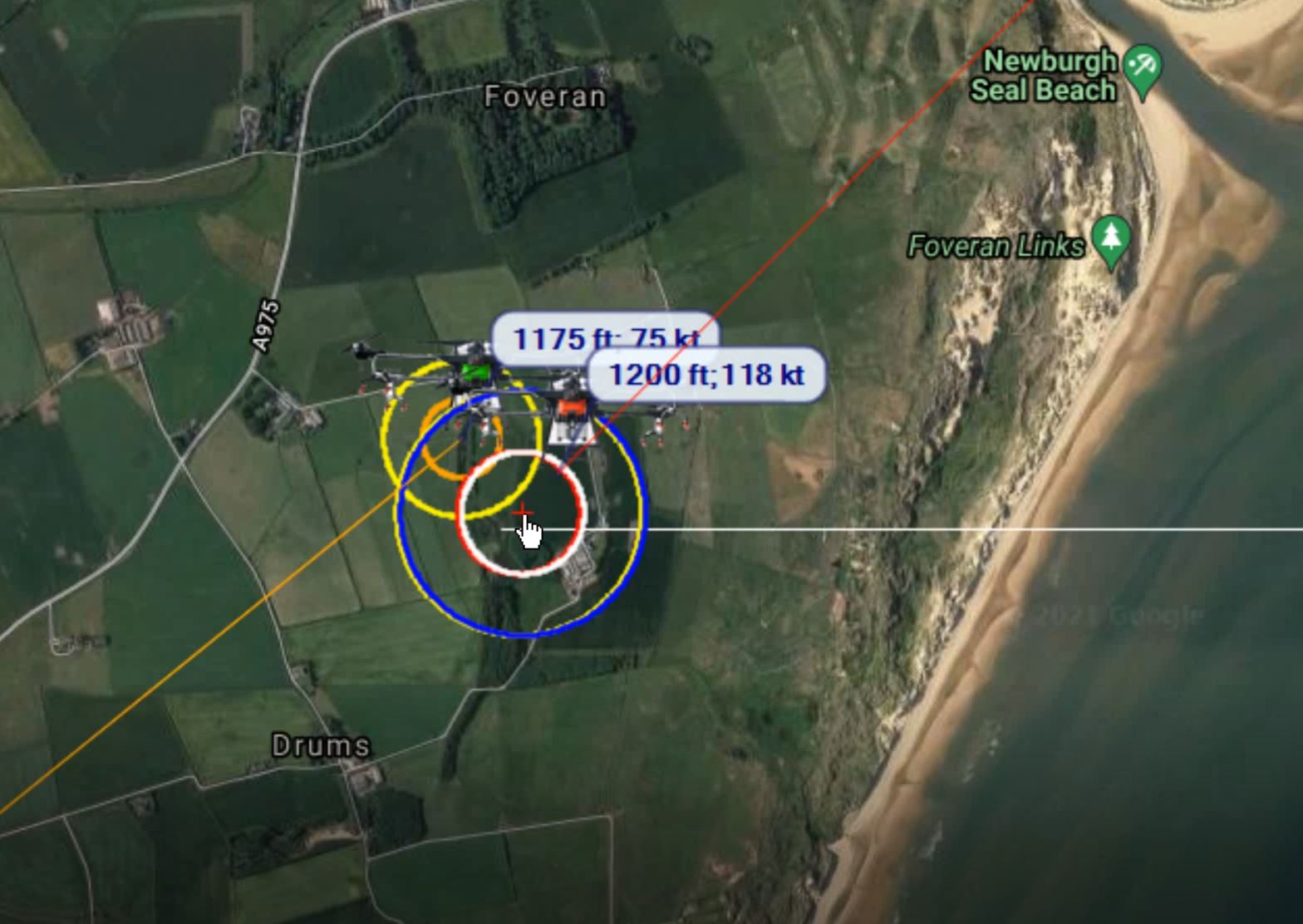}
	\end{center}
	\caption{Self manoeuvring using co-simulation -- simulated and real flight: Red $ICTZone$ and heading and yellow $PCTZone$ within the original path; white $ICTZone$ and heading and blue $PCTZone$ within the manoeuvring trajectory.}
	\label{fig:selfMan}
\end{figure}

$RoI$, in particular, $cFz$, is used optimally by considering the trajectories of other flights while planning autonomous manoeuvres without creating new collision risks during manoeuvres. $meD$ with a minimum manoeuvre strategy changes its altitudes either upward or downward on the vertical plane along with its direction on the horizontal plane without large deviation from the original track by taking $ICTZones$ of $meD$ and other flights, and velocity vectors of $meD$ and the flights leading to $IcRz$ into account. This minimum manoeuvre strategy with various generated waypoints resulting from the determined CA velocity vectors of $meD$ aims to get out of $ICTZones$ of other flights as soon as possible by avoiding any $IcRz$ with other flights in the vicinity. A successive course of autonomous manoeuvres are executed using an AI-based diversion strategy until $meD$ safely avoids the confronted $IcRz$ considering the accurate calculations of the instant angular velocity of other flights as illustrated in Fig.~\ref{fig:AICA}, formulated in Table~\ref{eqMan} and shown in Fig.~\ref{fig:selfMan} with white $ICTZone$ and heading and blue $PCTZone$. The radius of the white sphere ($ICTZone$) is 182 m with 2.5 sec TD and it is 364 m for the blue sphere ($PCTZone$) with 5 sec TD where the speed is 118 kn regarding $meD$. The radius of the orange sphere ($ICTZone$) is 116 m with 2.5 sec TD and it is 232 m for the yellow sphere ($PCTZone$) with 5 sec TD where the speed is 75 kn regarding the other aircraft. In other words, self manoeuvres start when the distance between the two flights is less than 238 m (182 + 116) in this case. This distance represents 5 sec TD that is corresponding to $IcRz$ between the two flights (i.e., $ICTD_{meD} + ICTD_{other}$) (Fig.~\ref{fig:ICTZones}). The minimum diversion manoeuvre can be noticed from the movement of the original red sphere to the white one leading to the self-separation of the two individual $ICTZones$ -- inner imminent collision spheres. Readers are referred to the recorded video entitled DACM-Sim-WithCA in the supplements for detailed manoeuvring actions in the various geospatial positions involving the flights in the same routing line going in the opposite direction. It is noteworthy to underline that the faster the speed the larger the drawn zones. The deconflict manoeuvre distance to the next waypoint is mainly determined considering the invasion of $ICTZones$ ($meD$.MoveDistance = $IcRz$\textsubscript{$RoI$.Flights(i)} - DirectDistance\textsubscript{$RoI$.Flights(i)} in Alg~\ref{alg:CA-AI}). In other words, how much the flights invade each other's $ICTZone$ is measured by corresponding to the minimal diversion distance to separate these zones from each other with minimal diversion manoeuvre. The altitude change (i.e., VerticalMv in Alg~\ref{alg:CA-AI}) regarding the manoeuvre distance of $meD$ is measured concerning the vertical manoeuvrability aerodynamic feature of $meD$ in metres specified in the properties of $meD$. The new heading of $meD$ (i.e., HeadingCA in Alg~\ref{alg:CA-AI}) for the manoeuvre leading to the instant separation of $meD$ from the other flight is determined regarding both the direction of the other flight and the horizontal manoeuvrability aerodynamic feature of $meD$ in degrees (i.e., maximum turning ability) specified in the properties of $meD$. The resulting trajectories from the next course of manoeuvres within the minimum manoeuvre strategy are composed of conflict-free waypoints. Warning messages are sent to the base stations (BSs) for the collision risks of FA-UAVs wherever possible. We improved the approaches regarding the results obtained from the real-world tests as explained earlier. During one of these tests, we noticed that very slow speeds can create collision risks where the $ICTZones$ of encountering aircraft get highly smaller and manoeuvring reaction may not cope with the limitations of the system (i.e., the total minimum response time, 2.5 sec) (elaborated in Section~\ref{sec:Limitations}). In this respect, it is worth noting that the aforementioned pilot and autonomous reaction time is halved regarding $IcRz$ and $PcRz$ if $meD$ or the other flight is hovering in the air with a steady position. Therefore, an emergency reaction distance can be identified to mitigate imminent collision risks caused by the very slow velocity of the flights. The system does not let the flights come closer than this distance by triggering autonomous manoeuvres as explained above.		

Imminent collision risks are capped within the horizontal and vertical planes and $meD$ returns to its original trajectory smoothly within the preplanned route after the collision is avoided with the aforementioned minimum manoeuvre strategy. This process is performed in parallel with the characteristics of the mission and the predefined/predetermined properties of $meD$ (Alg.~\ref{alg:CollisionAvoidance}). More specifically, $meD$ returns to the point where the CA manoeuvre has started if the mission type is ground-based (e.g., landmine detection, search and rescue) whereas it returns to the nearest point in the preplanned route if the mission type is air-based (e.g., logistics). The process of returning to the original route starts after $meD$ avoids $PcRz$ with the flight causing $IcRz$ for ground-based missions in order not to result in further $IcRz$ with the same flight as observed from the experiments.
 
\begin{figure}[htp]
	\begin{center}
		\includegraphics[width=0.31\textwidth]{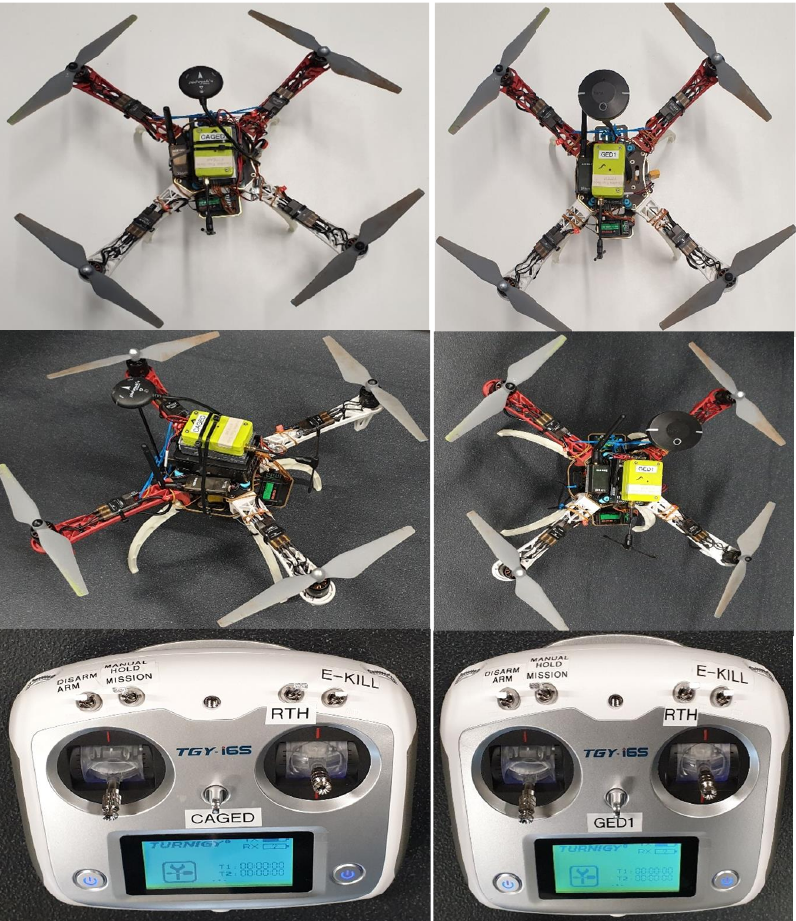}
	\end{center}
	\caption{Hardware components of the rotary drones developed to test the system. CA Grafiti Equipped Drone(Left): Durandal (PX4) AutoPilot, DJI F450 Quadcopter Frame, DJI 960kv Motors x 4, DJI 4S 20A ESC x 4, TGY-IA6B 2.4GHz Receiver, TGY-I6S RC Controller, Pixhawk4 GPS Module, Raspberry Pi 4 (Companion Computer). Grafiti Equipped Drone (right): PixHawk2 Hex Cube (PX4) AutoPilot, DJI F450 Quadcopter Frame, DJI 960kv Motors x 4, DJI 4S 20A ESC x 4, TGY-IA6B 2.4GHz Receiver, TGY-I6S RC Controller, Here GPS Module and PilotAware Grafiti EC module.}
	\label{fig:HWDrones}
\end{figure}

\subsubsection {Simulation module}
\label{sec:Simulation}
This module is designed to test and enhance the techniques developed in the system. In the module (Fig.~\ref{fig:methodology} F), many different types of scenarios involving real flight data can be generated with various conflicts and the techniques can be tested using those co-simulated scenarios before being deployed in real-world implementations. 

Simulated data that includes preplanned routes of $meD$ can be co-simulated with real-world flight data continuously (e.g., flights around an airport) leading to a diverse range of spontaneous varying scenarios with live flights. Technically speaking, improvised simulated data for $meD$ involving several other simulated flights and spontaneous real-world flight data are streamed into the system as if they are in the same environment to test how $meD$ detects and reacts to conflicts and collision risks in every possible situation using the developed modules and techniques within those modules (Fig.~\ref{fig:methodology}). These unplanned scenarios can be saved and rerun with the techniques repetitive times until the desired outcome is obtained while the techniques are being improved considering the existing unsatisfactory results. In this regard, the developed techniques, in particular, the CA techniques (Algs.~\ref{alg:CollisionAvoidance} and~\ref{alg:CA-AI}) were advanced with many spontaneously generated scenarios. 

\section {Experimental Design} 
\label{sec:Design}

\subsection {Environmental settings}
\label{sec:HW}
\begin{figure}[htp]	
	\begin{center}
		\includegraphics[width=0.36\textwidth]{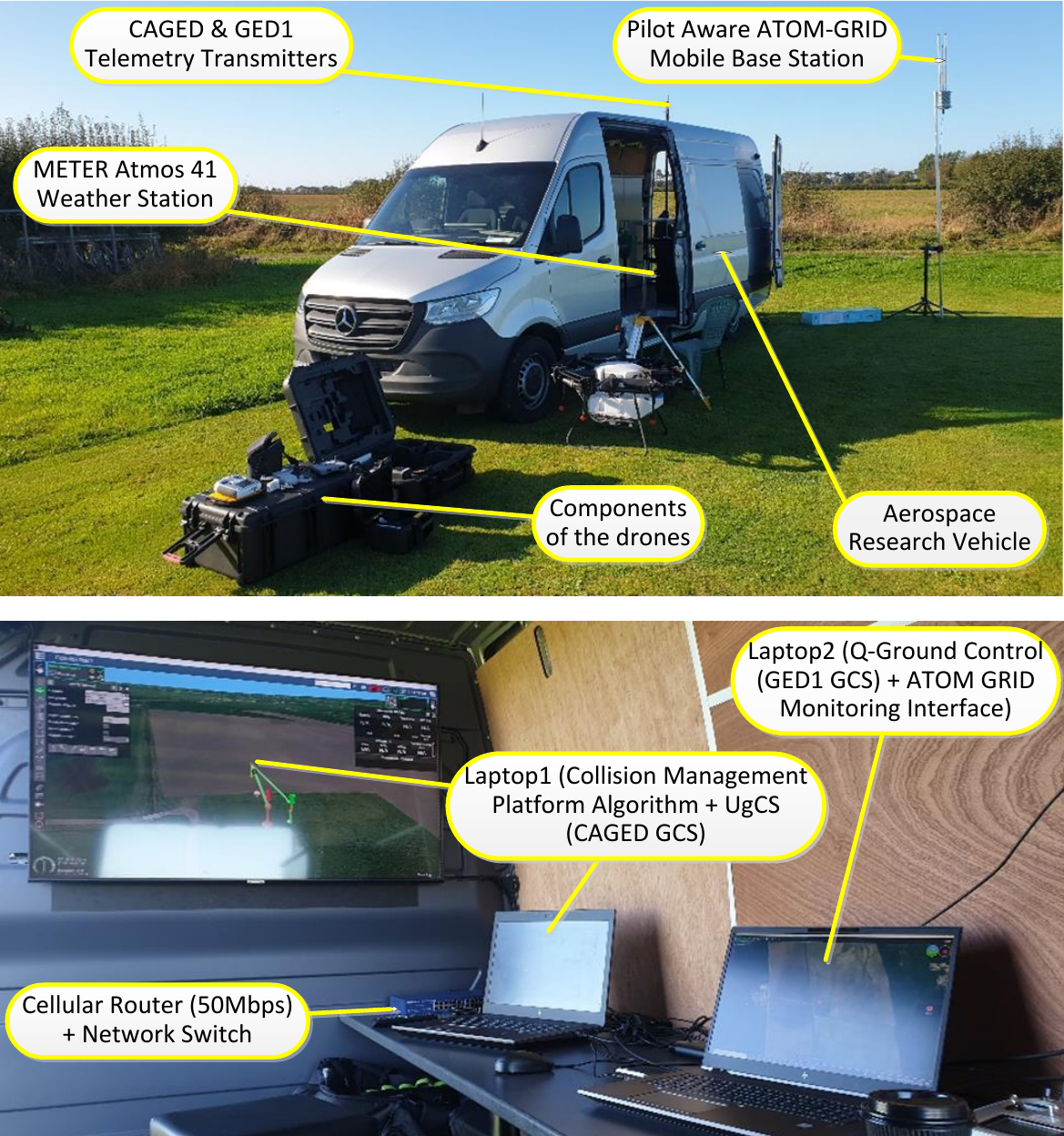}
	\end{center}
	\caption{Settings with UCLan aerospace VAN.}
	\label{fig:Settings}
	
\end{figure}

The drones, developed by our team, with the purpose of testing the DACM system, are equipped with the PilotAware EC device as depicted
in Fig.~\ref{fig:HWDrones} with their properties.
These drones are capable of transmitting their flight data using these onboard EC devices as explained in Section~\ref{sec:filtering}. $PCTD$ and $ICTD$ are defined as 5 sec and 2.5 sec $TD$ respectively whereas the overlapping $PCTZones$ generate $PcRz$ with 12 sec $TD$ as formulated in Table~\ref{eq1} and the overlapping $ICTZones$ generate $IcRz$ with 5 sec $TD$ as formulated in Table~\ref{eq23} in our experiments as explained in Section~\ref{sec:mapOperations}. The environmental settings of the system are presented in Fig.\ref{fig:Settings}. The computational tasks are offloaded to the ground-centralised nodes with these settings to extend the flying time of drones where all the computation can be processed using the onboard companion Raspberry Pi 4, which may be feasible for fixed-wing UAVs as clarified in Section~\ref{sec:Techniques}. The improvement of the techniques was carried out through a cycle of various simulation tests as explained in Section~\ref{sec:simulation} and real-world implementations as explicated in Section~\ref{sec:realTests}. One of the detailed test plans (58 slides (``CA Flight Test Plan v4.pdf''')) prepared by the team to perform them in both simulation and real-world environments is placed in the supplements. This detailed test plan can be utilised by the relevant research community for other similar studies.

\subsection {Test in simulation \& improvements of techniques}
\label{sec:simulation}
Simulated flight data is incorporated into live real-data using the simulation module (Section~\ref{sec:Simulation}) for i) co-simulation that can generate a limitless number of scenarios, ii) testing the techniques using these many numbers of scenarios before real-world trials to verify their efficacy under any possible circumstances and iii) improving the techniques if the desired results are not obtained or collisions are occurring during simulation tests. In this direction, the robustness of the developed system was ensured using hundreds of tests with varying improvised, i.e., uncontrolled or no pre-planned scenarios. Interested readers are referred to the 25-sec videos in the supplements, "DACM-Sim-NoCA" with no CA manoeuvres and "DACM-Sim-WithCA" with CA manoeuvres recorded during one of these tests for observing distinctively how the techniques are executed with the 3D geospatial manoeuvres where the mission type is air-based.

\begin{figure}[htp]
	\begin{center}
		\includegraphics[width=0.48\textwidth]{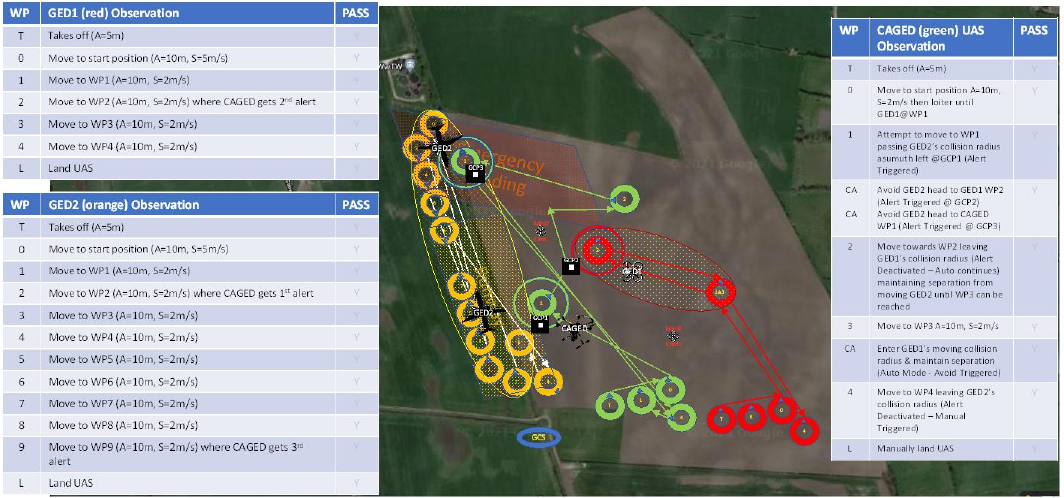}
	\end{center}
	\caption{Phase 9: Test 4 - Avoid multiple moving collisions tracks for heterogeneous UAVs. Outbound track collides with GED2 (red) then collides with GED1 (yellow) then heads left towards GED2 maintaining parallel separation. Inbound track avoids collision flying parallel separation to a UAV. 
	}
	\label{fig:3DroneTest}
\end{figure}

\begin{figure*}[htp]
	\begin{center}
		\includegraphics[width=0.99\textwidth]{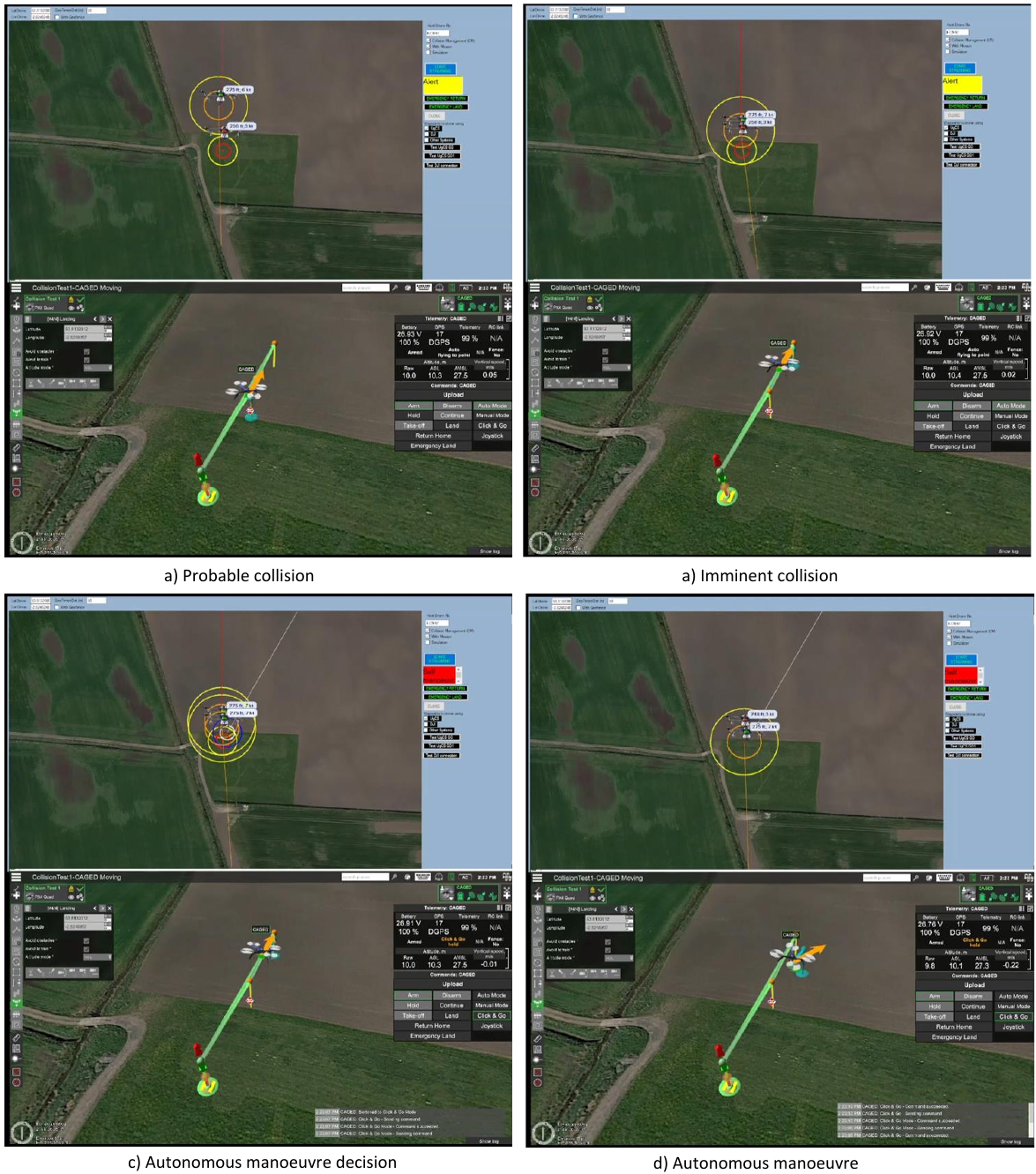}
	\end{center}
	\caption{Real field tests: Top images show the screenshots of the developed application whereas the bottom images show the UgCS control system at a time where the UgCS system is responsible for executing the modified control commands. The red circles indicate the spherical $ICTZones$ of $meD$ and the brown circles indicate the $ICTZones$ of the other flights in the environment; the yellow circles correspond to the spherical $PCTZones$ of flights. The circles for the autonomous successive course of manoeuvres of $meD$ turn into white $ICTZone$ and heading and blue $PCTZone$ as shown in c and d.}
	\label{fig:realTest}
\end{figure*}

\subsection {Test in real-world \& improvements of techniques}
\label{sec:realTests}
The safe, efficient and autonomous BVLOS operations of drones in dynamic aerial environments
are dependent on the abilities of drones -- i.e., `level of autonomy`, `sense', `SSA`, `manoeuvrability', `collision avoidance', `communication', `swarm intelligence', and `decision making' which are catered for in our proposed solution by adopting a modular and open software standard designed for interoperability and compatibility with a large variety of different drones. Five drones (1 fixed-wing and 4 rotaries) were built by the team. These drones (Fig.~\ref{fig:HWDrones}) along with manned aeroplanes were deployed to test and improve the developed techniques in real-world scenarios in addition to many co-simulated scenarios (Section~\ref{sec:simulation}). 

\begin{figure}[htp]
	\begin{center}
		\includegraphics[width=0.4\textwidth]{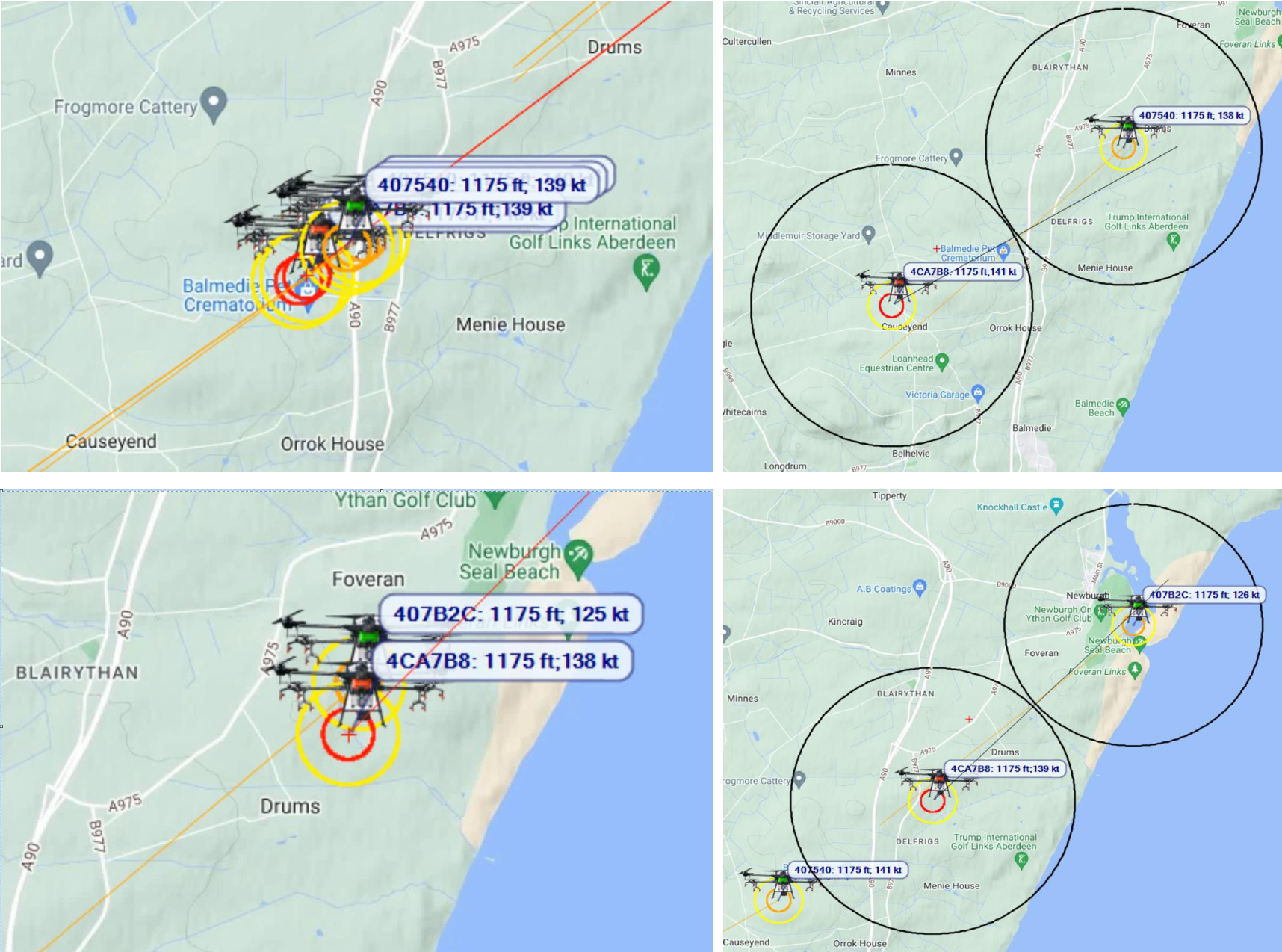}
	\end{center}
	\caption{Deviation moments (left) of the aircraft from original trajectories during collision avoidance manoeuvrers at the same altitude and MAC avoidance standard (black circles) with 30 sec radius (right) for these encounters. The aircraft with the ID, 4CA7B8 is co-simulated whereas the other aircraft with the IDs, 407540 and 407B2C, are real flights.}
	\label{fig:Deviation}
\end{figure}

The real-world implementation of the system was fulfilled at two locations with the coordinates of 53.813222, -2.824523 and 53.923917, -0.977914 where UAV flight tests are allowed by the local government. Five field tests involving multiple scenarios were conducted. The first test with 2 drones and the manned aeroplane, piloted by Keith Vinning within the team, accompanied by the other two manned aeroplanes was carried out to analyse the infrastructure, the performance of data sharing between the BSs and data streaming abilities between vehicles and BSs. 
The core techniques have been developed mainly based on the data obtained from this first real-world experiment. A part of the recorded test video along with the test plan (4 slides (`First test plan (aeroplanes and UAVs).pdf')) is in the supplements. An example of further tests is shown in Fig.~\ref{fig:realTest} using two drones for the sake of simplicity to observe how the developed system can operate in real-world encounters. The test video recorded with three drones is in the supplements. One of the detailed test plans (``CA Flight Test Plan v4.pdf''') involving more complex real-world scenarios is also placed in the supplements. For instance, the test scenario with multiple heterogeneous UAVs is presented in Fig.~\ref{fig:3DroneTest}. The achieved functionalities of the system were demonstrated on multi-rotors whose properties are presented in Fig.~\ref{fig:HWDrones}. A video (titled ``FA-UAV-encounters-noncooperative-drone.MP4'') recorded by the camera mounted on a third drone is in the supplements to demonstrate how a FA-UAV within its preplanned route is avoiding a collision with a non-cooperative drone and returning its preplanned trajectory while showing how the interface of the DACM system is running in parallel with the manoeuvres actuated for collision avoidance and returning to the preplanned route. Incorporation of fixed-wing drones into more complex scenarios has been planned as a future work as pointed out in Section~\ref{sec:conclusion}. 

The geofencing module (Fig.~\ref{fig:methodology} E) is autonomously engaged to cancel the autonomous manoeuvres that are taking place out of the predefined restricted region to mitigate the risks of unaccepted behaviours of $meD$. The outcomes of the previous tests fed the following test results and helped improve the techniques further. 

\section{Experimental Results}
\label{sec:Results}
The real-world and co-simulated experiments were performed using the minimum TD reaction time, i.e., 2.5 sec and 5 sec $TD$ reaction time for the $ICTZones$ and $PCTZones$ (Figs.~\ref{fig:PCTZones} and~\ref{fig:ICTZones}) respectively concerning the limitations of the developed system (i.e., the total minimum response time, 2.5 sec) (elaborated in Section~\ref{sec:Limitations}). Time-based reaction $TD$ produces changing radius for the $ICTZones$ and $PCTZones$ based on the speed of the aircraft. The autonomous manoeuvrers were actuated 5 sec (2.5 sec for the $ICTZone$ of the first aircraft + 2.5 sec for the $ICTZone$ of the second aircraft) before an imminent collision, which allowed the autonomous drones in the experiments to have sufficient time for avoiding any imminent collisions safely. Manoeuvring took place concerning the borders of the $ICTZones$ where the system aims to separate the two borders of the $ICTZones$ of the aircraft during a collision risk, but not further to provide the minimum deviation strategy that helped the drone in manoeuvring return to the original trajectory as efficiently as possible to complete its mission as planned. For instance, the additional distance to the preplanned route of the aircraft with the ID, 4CA7B8 was 408.12 metres during autonomous manoeuvres to avoid collision risks with the aircraft whose ID is 407540 at the top image in Fig.\ref{fig:Deviation} considering the deviation manoeuvres from the original trajectory whereas it was 322.25 metres in the bottom image with the aircraft whose ID is 407B2C while the radius of the $ICTZones$ for the manoeuvring aircraft (i.e., 4CA7B8) was 178.77 metres with a speed of 138 kt. It is noteworthy to emphasise that the additional distance to the preplanned route is less for the bottom image compared to the top image where i) the encounter at the top image is severe, ii) the speed of the other aircraft (i.e., 125 kn for 407B2C) at the bottom image is less than the other aircraft at the top image (i.e., 138 kn for 407540), producing a smaller radius of the $ICTZones$, and iii) larger radius of the $ICTZones$ at the top image necessitate a larger deviation. It is observed in real-world tests that the more the speed of flights in collision risks where the $ICTZones$ get larger regarding the higher velocities, the more the acceleration of the aircraft to leave the $ICTZones$ to reach its determined manoeuvring point.

Experimental results demonstrated that collision avoidance strategy can be standardised with centimetre-based anticipation manoeuvrers for aircraft to avoid mid-air collisions, which cannot be realised using the onboard sensors with multiple shortcomings considering the MAC distance standards (Fig.\ref{fig:Deviation}). The results proved the robustness of the CM techniques developed in this study. The objectives of the project have been achieved by forging various expertise from the industry and university within a well-established multi-disciplinary team. 
$meD$ can manoeuvre autonomously to avoid collisions without diverting its path significantly if its spherical $ICTZone$ is violated by other flights` spherical $ICTZones$ within $IcRz$ and it can return to its preplanned route smoothly. The main achieved objectives of the DACM system can be summarised in three categories as follows:

1) Sense and Alert: The drone can detect flights in a low collision risk range and alert the pilot about them. More specifically, the DACM system coordinates the probable actions of PC-UAVs with their pilots first by raising a collision awareness where there is a probable collision risk (i.e., no immediate danger of a collision) within $PcRz$. This is possible from Level-1 of drone autonomy onwards.

2) Sense, Alert and Avoid: The drone can detect flights in a low collision risk range, alert the pilot about them with a probable collision awareness within $PcRz$ and react autonomously to avoid imminent collision risks within $IcRz$ in a high collision risk range independent from the previous decision of the pilot regarding the probable collision awareness. This is possible from Level-3 onwards. 

3) Sense, Avoid and Navigate: The drone can detect flights in a high collision risk range, divert its navigation path autonomously for avoiding imminent collision risks within $IcRz$ and return to its preplanned route, smoothly in an autonomous manner after the collision is avoided. In this category, the fully autonomous mission planning of the drones is aimed with necessary manoeuvres regarding the imminent collision risks. This is only possible from Level-4 of drone autonomy by taking the human out of the loop, which is achieved to allow for autonomous optimised navigation by keeping the drone well clear of other traffic within its predefined route.

\section{Discussion}
\label{sec:discussion}
NATS has confirmed that, from their data interpretation, over 1,000 airspace infringements on aerospace law occur each year; this figure sets infringements as one of the highest-risk events that NATS face at this time; As traffic levels increase, it is important that we recognise the more likely need for focused intervention by all stakeholders, to ensure that the potential for mid-air collision does not increase by pursuing and encouraging targeted and continuous improvements in technology and their respective systems~\cite{UKCAA23}. Swarm intelligence algorithms have been utilized in numerous diverse domains for solving optimization problems, e.g., scheduling problems, robots, power systems, parameter optimization, system identification, image processing, and signal processing~\cite{9498989}. Swarms of UAVs equipped with a variety of sensors are commonly deployed to realise a diverse range of missions along with in-situ sensing faster and efficiently due to their flexible use with excellent manoeuvrability~\cite{8611082}. The probability of collision risks posed by UAVs is increasing as public use of drones continues to grow and autonomous commercial applications are now becoming a reality. The proposed technology is expected to contribute to swarm intelligence algorithms in solving optimization problems while increasing task performance and safety. In our other funded research, with the same team, we have analysed swarm intelligence with extensive experiments while performing assigned tasks using the technology elaborated in this research with multiple autonomous UAVs within human-robot co-work dialogue~\cite{Kkuru24}. Autonomy is defined as the ability of a system to sense, communicate, plan, make decisions and act without human intervention~\cite{YBSebbane15}. 
FA-UAVs as flying autonomous robots determining their course of action with onboard sensor data analysis involving autonomous take-off and landing are taking their place in real-life to accomplish many different tasks~\cite{9314128}. Due to the limitation of payloads, it is infeasible to carry sophisticated heavy sensors, which would lead to increasing power consumption and drastically decrease the flight time~\cite{8494201}. Moreover, due to the limited power available onboard, UAVs must make careful decisions about how to best utilise power for communication~\cite{doi:10.2514/6.2012-2455} and processing where they both consume high amounts of energy. This paper, aiming both to mitigate the aforementioned concerns and to provide an effective CM system, realises the safe use of FA-UAVs by keeping each UAV well clear of other traffic using effective DAA techniques by which an optimal CA strategy, resulting in minimum deviations from the original trajectory, is implemented based on the relative geometric features of close-range flights.

Full automation is anticipated to become a reality in the near future and will bring significant benefits, particularly to the transport and logistics sector. AI will be the key enabler in improving the autonomy of UAV applications. Most AI techniques such as deep neural networks (DNN) requiring prior knowledge and deep reinforcement learning (DRL) requiring less prior knowledge aim to solve the classification problem. Recently, DRL, using Q-learning, has been immensely employed to solve the path planning problem and maze problems of robots. However, these approaches, suffering from high variance and low reproducibility, are not designed to solve the instant CA problem of UAVs under highly dynamic conditions with many uncertainties where RL does not have a fixed training set with ground-truth labels~\cite{9001167}. More specifically, the main goal of Q-learning is to train a general model for the same set of tasks. Q learning algorithm has the problem of a low reuse rate, and different environments need to be re-trained, so reducing the training time is the key to the implementation of RL. Machine learning algorithm has high computational complexity, high resource overhead and dependence on a large number of prior data, which makes it difficult to deal with dynamic and unknown scenarios~\cite{9181515}. In this sense, a computationally less intensive multi-agents explainable AI (XAI) methodology considering physical constraints, environmental constraints and current specific characteristics (e.g., velocity vectors) of flights is incorporated into the developed DACM system to build an effective CM policy requiring no prior knowledge and training. Using real-time data provided by the PilotAware system (Table~\ref{tab:DataStreaming}) including identification, position, speed, heading and altitude, it has been possible to determine if the intruder aircraft is increasing, reducing or maintaining a level altitude with varying velocity vectors, therefore various avoid tactics using AI techniques can then be planned. These can either inform the drone pilot for approval or be autonomously executed depending upon the level of risk determined by the algorithm or operator. 

Compared with a single platform, cooperative autonomous UAVs offer efficiency and robustness in performing complex tasks~\cite{4967999}. In this sense, the development of effective CA systems is of great significance to performing collective and cooperative tasks, in particular, using the global coverage strategy leading to an excellent SA. Many drone applications can benefit from a unified framework that coordinates their access to aerospace and helps them navigate to the points of interest where they have to perform a task~\cite{7423671}. Any architecture poised to provide this service must be scalable and be able to provide it to thousands of drones, which will share the congested and limited urban aerospace~\cite{7423671}. It is a high priority to deploy FA-UAVs in optimised routes by taking the non-autonomous UAVs and manned aircraft into consideration involving the city structures, facilities and weather conditions to accomplish coordinated and cooperative missions safely. National and supranational authorities (e.g., the Federal Aviation Administration (FAA), European Union Aviation Safety Agency (EASA), and International Civil Aviation Organisation) and industry actors (such as Amazon, Google, and drone manufacturer DJI) are now developing systems for UTM~\cite{9061133}. The main aim of UTM is to achieve safe and efficient UAV operations~\cite{9061133}. The safe integration of UTM systems into ATM systems using intelligent autonomous approaches is an emerging requirement to manage the highly dynamic shared aerospace where the number of wide-scale deployments of public and commercial highly heterogeneous UAV applications is steadily growing owing to their flexible and cost-effective use. This study aims to enable aircraft and drones to share the same aerospace without segregation where there are plans to make EC devices mandatory in UAVs for effective air traffic management in the USA, EU and other parts of the world~\cite{8396612}. The primary motivation behind this study is to help build trust for the use of autonomous UAVs. DACM is expected to be utilised in increasing the efficacy of the UTM system and its integration with the ATM system to autonomously cope with distributed multiple encounters and handle dynamic path planning in integrated aerospace. The viability of the DACM system and the results obtained from the co-simulated and real-world tests suggest that not only can UAVs be safely integrated into ATM systems, but also the standardised size of the MAC avoidance borders as exemplified with the right images in Fig.~\ref{fig:Deviation} (Section~\ref{sec:Literature}) can be reduced significantly without compromising safety to increase the efficient use of aerospace with the deployment of the intelligent systems similar to the proposed system in this research.

\section {Conclusion and Future Work}
\label{sec:conclusion}
Intelligence sources continue to show that MAC remains a key risk in airspace because of the busy and complex airspace serving the individual needs of commercial, military, general aviation and `new entrant' airspace users~\cite{UKCAA23}. Establishing a high level of trust among regulatory organizations, the public, and the industry is paramount to deploying self-operating UAVs effectively in many public and commercial applications leading to help build their game-changing role. Both integrating drones, particularly, autonomous UAS for BVLOS operations, into ATM systems safely and their efficient use as well as the efficient use of aerospace require the development of new technologies and approaches (as elaborated in Section~\ref{sec:Literature}), which rules out collision risks within ATM systems. In this context, this paper aims to establish the safe use of UAVs within UTM systems that can be fully integrated into ATM systems. Autonomous air traffic management by avoiding any collision is targeted by the aviation authorities whereas full autonomy for UAVs by removing the remote pilot from the loop is the ultimate goal in scaling commercial UAVs operating BVLOS, not only from a conceptual point of view, but also from an economic point of view. In this paper, we have presented a time-optimal CM framework for UAVs by raising the SA of flights using a global coverage strategy equipped with the PilotAware grid system. An efficient CM methodology, the so-called DACM system that reduces the spatial deviation from the trajectory whilst keeping a safe distance from the other flights in the surrounding environment depending on the relative geospatial position and velocity vectors is developed. This system can satisfy the regulations of the aviation authorities while using drones in ATM systems concerning the definition of a MAC that is likely within a 30-60 seconds time window~\cite{doi:10.2514/1.D0091}, whereas the aforementioned drone sensor technologies (Section~\ref{sec:Literature}) with various shortcomings cannot ensure this under dynamic environmental conditions where the flights with high velocities is not within range of the UAS sensors regarding a 30-60 seconds separation. The developed technology works independently from the environmental constraints and outperforms the current onboard UAS sensor systems in avoiding MAC. Simulation and experimental real-field flight tests, which were performed within 2.5-sec proximity to manoeuvre, demonstrate an effective proof-of-concept system that can be used in more complex environments putting heterogeneous UAVs with different flight characteristics to use. The aviation rules can be mandated considering the accurate measurement abilities provided by the proposed system, which cannot be guaranteed using the aforementioned onboard drone sensor technologies. Most importantly, a 2.5-sec successful manoeuvring ability that enables avoiding any MAC allows much smaller time frames than 30-60 seconds (e.g., < 10 sec). This facilitates a safer and more efficient air traffic system with mixed traffic with heterogeneous flight characteristics. The results obtained from those tests verify that collision-free trajectory for both PC- and FA-UAVs is assured with the proposed approach without sophisticated onboard sensors and prior training. The developed approach has made it possible to prevent MAC thoroughly, especially between high-velocity flights, which are currently unavoidable due to the limitations of off-the-shelf sensor technologies. This can be achieved by implementing a global coverage strategy that utilises EC information obtained from all flights, without introducing new collision risks, which allows precise measurements with excellent state and situation awareness. DACM, with an effective DAA and low computational requirements enabling UAVs, particularly, FA-UAVs, to remain well clear from other UAVs and manned aeroplanes, is expected to be advised by the aviation regulatory organisations in order to provide safe air space involving manned aircraft, in particular, during the autonomous BVLOS operations, which will pave the way for the deployment of FA-UAVs on a large scale in dense air traffic environments for completing swarms of multiple complex diverse missions simultaneously. 

The future works can be summarised as follows: i) incorporate the fixed-wing drone (presented in footnote$^1$) developed by the team into further test scenarios as a host drone ($meD$) after its ongoing functional tests have been completed, ii) port the developed algorithm onto an onboard companion computer should any drone lose communication or any fault occur with the ground control station, iii) test the onboard CM system in dense traffic using larger heterogeneous types of UAVs, and finally iv) UCLan will provide access to its investment readiness programme to help commercialise the technology.



\section {Limitations of DACM System }
\label{sec:Limitations}
The identification of $ICTD$ has to be larger than 2.5 sec $TD$ to avoid any imminent collisions safely where 1 sec is for the delay of the flight data transfer from the PilotAware system, 0.5 sec (i.e., around \%25 for the module of ``B. Map operations: Geospatial error correction'', \%70 for the module of ``B. Map operations: MoI \& RoI geospatial/geometric operations'' and \%5 for "Finding the most appropriate manoeuvre" in module D (Fig.~\ref{fig:methodology}) is for the processing of the data to determine the right course of action for 10 miles ROI with SA of 10 flights as the worst-case scenario; the smaller the ROI with a reduced number of flights, the less the processing time. 1 sec is required to let the drone system take the determined desired action and this time will be reduced significantly with the onboard computing (item ii in future works above). The total minimum response time, 2.5 sec, is aimed to be reduced to under 1.5 sec in the further design of the system within a newly developed infrastructure by PilotAware Ltd (https://www.youtube.com/watch?v=JCpXdSFtHmU).


\section* {Funding Information and Acknowledgements}
\label{sec:Funding}
This report is independent research funded by the Department for Transport (DfT) (Drones-Technology Research Innovation Grants (D-TRIG) 2021 Programme) $^1$. The funding agreement ensured the authors' independence in designing the study, interpreting the data, writing, and publishing the report. The views expressed are those of the authors and not necessarily those of DfT. We would like to express our gratitude for this fund. Besides, the authors would like to thank the anonymous reviewers for their constructive comments.

\bibliographystyle{IEEEtran}
\bibliography{vehiclesUAV}	

\begin{thebibliography}{10}
\providecommand{\url}[1]{#1}
\csname url@samestyle\endcsname
\providecommand{\newblock}{\relax}
\providecommand{\bibinfo}[2]{#2}
\providecommand{\BIBentrySTDinterwordspacing}{\spaceskip=0pt\relax}
\providecommand{\BIBentryALTinterwordstretchfactor}{4}
\providecommand{\BIBentryALTinterwordspacing}{\spaceskip=\fontdimen2\font plus
\BIBentryALTinterwordstretchfactor\fontdimen3\font minus
  \fontdimen4\font\relax}
\providecommand{\BIBforeignlanguage}[2]{{%
\expandafter\ifx\csname l@#1\endcsname\relax
\typeout{** WARNING: IEEEtran.bst: No hyphenation pattern has been}%
\typeout{** loaded for the language `#1'. Using the pattern for}%
\typeout{** the default language instead.}%
\else
\language=\csname l@#1\endcsname
\fi
#2}}
\providecommand{\BIBdecl}{\relax}
\BIBdecl

\bibitem{4337970}
S.~{Bouabdallah}, M.~{Becker}, and R.~{Siegwart}, ``Autonomous miniature flying
  robots: coming soon! - research, development, and results,'' \emph{IEEE
  Robotics Automation Magazine}, vol.~14, no.~3, pp. 88--98, 2007.

\bibitem{MRadovic19}
M.~Radovic, ``Tech talk: Untangling the 5 levels of drone autonomy,'' 2019.

\bibitem{doi:10.1002/tee.23041}
K.~Nonami, ``Present state and future prospect of autonomous control technology
  for industrial drones,'' \emph{IEEJ Transactions on Electrical and Electronic
  Engineering}, vol.~15, no.~1, pp. 6--11, 2020.

\bibitem{9001167}
D.~Wang, T.~Fan, T.~Han, and J.~Pan, ``A two-stage reinforcement learning
  approach for multi-uav collision avoidance under imperfect sensing,''
  \emph{IEEE Robot Autom Lett}, vol.~5, no.~2, pp. 3098--3105, 2020.

\bibitem{9847033}
J.~Tang, X.~Chen, X.~Zhu, and F.~Zhu, ``Dynamic reallocation model of multiple
  unmanned aerial vehicle tasks in emergent adjustment scenarios,'' \emph{IEEE
  Trans. Aerosp. Electron. Syst.}, vol.~59, no.~2, pp. 1139--1155, 2023.

\bibitem{9256553}
M.~Pálenská, S.~L. Brázdilová, P.~Cásek, and L.~Korenčiak, ``Low-power
  ads-b for ga operating in low altitude airspace,'' in \emph{2020 AIAA/IEEE
  39th Digital Avionics Systems Conference (DASC)}, 2020, pp. 1--10.

\bibitem{9314128}
K.~{Kuru}, ``Planning the future of smart cities with swarms of fully
  autonomous unmanned aerial vehicles using a novel framework,'' \emph{IEEE
  Access}, vol.~9, pp. 6571--6595, 2021.

\bibitem{8379533}
J.~{Boubeta-Puig}, E.~{Moguel}, F.~{Sánchez-Figueroa}, J.~{Hernández}, and
  J.~{Carlos Preciado}, ``An autonomous uav architecture for remote sensing and
  intelligent decision-making,'' \emph{IEEE Internet Comput.}, vol.~22, no.~3,
  pp. 6--15, 2018.

\bibitem{8573148}
J.~Zhang, J.~Yan, and P.~Zhang, ``Fixed-wing uav formation control design with
  collision avoidance based on an improved artificial potential field,''
  \emph{IEEE Access}, vol.~6, pp. 78\,342--78\,351, 2018.

\bibitem{8675275}
K.~{Kuru} and H.~{Yetgin}, ``Transformation to advanced mechatronics systems
  within new industrial revolution: A novel framework in automation of
  everything (aoe),'' \emph{IEEE Access}, vol.~7, pp. 41\,395--41\,415, 2019.

\bibitem{ViqueratAD2008RCAf}
A.~Viquerat, L.~Blackhall, A.~Reid, S.~Sukkarieh, and G.~Brooker, ``Reactive
  collision avoidance for unmanned aerial vehicles using doppler radar,'' pp.
  245 -- 254, 2008.

\bibitem{8396612}
S.~Chindea, P.~Iravani, J.~Luke~du Bois, D.~Cleaver, and A.~Lawrenson,
  ``Holistic approach to safe operation of small uas in class g airspace,'' in
  \emph{2018 IEEE Aerospace Conference}, 2018, pp. 1--12.

\bibitem{8684792}
R.~P. Padhy, S.~K. Choudhury, P.~K. Sa, and S.~Bakshi, ``Obstacle avoidance for
  unmanned aerial vehicles: Using visual features in unknown environments,''
  \emph{IEEE Consum Electron Mag}, vol.~8, no.~3, pp. 74--80, 2019.

\bibitem{doi:10.1080/10095020.2017.1420509}
Y.~Lu, Z.~Xue, G.-S. Xia, and L.~Zhang, ``A survey on vision-based uav
  navigation,'' \emph{Geo Spat Inf Sci}, vol.~21, no.~1, pp. 21--32, 2018.

\bibitem{7535138}
Y.~Lyu, Q.~Pan, C.~Zhao, Y.~Zhang, and J.~Hu, ``Vision-based uav collision
  avoidance with 2d dynamic safety envelope,'' \emph{IEEE Aerospace and
  Electronic Systems Magazine}, vol.~31, no.~7, pp. 16--26, 2016.

\bibitem{8611082}
K.~{Kuru}, D.~{Ansell}, W.~{Khan}, and H.~{Yetgin}, ``Analysis and optimization
  of unmanned aerial vehicle swarms in logistics: An intelligent delivery
  platform,'' \emph{IEEE Access}, vol.~7, pp. 15\,804--31, 2019.

\bibitem{8767930}
S.~H. Arul, A.~J. Sathyamoorthy, S.~Patel, M.~Otte, H.~Xu, M.~C. Lin, and
  D.~Manocha, ``Lswarm: Efficient collision avoidance for large swarms with
  coverage constraints in complex urban scenes,'' \emph{IEEE Robot Autom Lett},
  vol.~4, no.~4, pp. 3940--3947, 2019.

\bibitem{2015JGCD...38.1140J}
Y.~I. {Jenie}, E.-J.~v. {Kampen}, C.~C. {de Visser}, J.~{Ellerbroek}, and J.~M.
  {Hoekstra}, ``{Selective Velocity Obstacle Method for Deconflicting Maneuvers
  Applied to Unmanned Aerial Vehicles},'' \emph{J Guid Control Dyn}, vol.~38,
  no.~6, pp. 1140--1146, Jun. 2015.

\bibitem{9145644}
B.~Lindqvist, S.~S. Mansouri, A.-a. Agha-mohammadi, and G.~Nikolakopoulos,
  ``Nonlinear mpc for collision avoidance and control of uavs with dynamic
  obstacles,'' \emph{IEEE Robot Autom Lett}, vol.~5, no.~4, pp. 6001--8, 2020.

\bibitem{doi:10.2514/1.G001715}
Y.~I. Jenie, E.-J. van Kampen, C.~C. de~Visser, J.~Ellerbroek, and J.~M.
  Hoekstra, ``Three-dimensional velocity obstacle method for uncoordinated
  avoidance maneuvers of unmanned aerial vehicles,'' \emph{J Guid Control Dyn},
  vol.~39, no.~10, pp. 2312--23, 2016.

\bibitem{8536788}
Y.~Xiuxia, Z.~Yi, and Z.~Weiwei, ``Obstacle avoidance method of
  three-dimensional obstacle spherical cap,'' \emph{Journal of Systems
  Engineering and Electronics}, vol.~29, no.~5, pp. 1058--1068, 2018.

\bibitem{7947166}
J.~Seo, Y.~Kim, S.~Kim, and A.~Tsourdos, ``Collision avoidance strategies for
  unmanned aerial vehicles in formation flight,'' \emph{IEEE Transactions on
  Aerospace and Electronic Systems}, vol.~53, no.~6, pp. 2718--34, 2017.

\bibitem{doi:10.2514/1.G002607}
J.~Yang, D.~Yin, and L.~Shen, ``Reciprocal geometric conflict resolution on
  unmanned aerial vehicles by heading control,'' \emph{J Guid Control Dyn},
  vol.~40, no.~10, pp. 2511--23, 2017.

\bibitem{lebron1983system}
J.~E. Lebron, A.~Zeitlin, N.~Spencer, J.~Andrews, and W.~Harman, ``System
  safety study of minimum tcas ii (traffic alert and collision avoidance
  system),'' \emph{MITRE, Technical Rep. MTR-83W241}, 1983.

\bibitem{doi:10.2514/1.D0091}
A.~Weinert, S.~Campbell, A.~Vela, D.~Schuldt, and J.~Kurucar, ``Well-clear
  recommendation for small unmanned aircraft systems based on unmitigated
  collision risk,'' \emph{Journal of Air Transportation}, vol.~26, no.~3, pp.
  113--122, 2018.

\bibitem{doi:10.2514/6.2010-9333}
\BIBentryALTinterwordspacing
M.~Kochenderfer, D.~Griffith, and J.~Olszta, \emph{On Estimating Mid-Air
  Collision Risk}. [Online]. Available:
  \url{https://arc.aiaa.org/doi/abs/10.2514/6.2010-9333}
\BIBentrySTDinterwordspacing

\bibitem{pointon2018integration}
J.~Pointon, \emph{Integration of NZDF Remotely Piloted Aircraft Systems (RPAS)
  Into New Zealand Civil Airspace}, ser. DTA report.\hskip 1em plus 0.5em minus
  0.4em\relax Defence Technology Agency, 2018.

\bibitem{USDoT11}
USDOT, ``Introduction to tcas ii,'' 2011.

\bibitem{9256776}
B.~Duffy, S.~Balachandran, A.~Peters, K.~Smalling, M.~Consiglio, L.~Glaab,
  A.~Moore, and C.~Muñoz, ``Onboard autonomous sense and avoid of
  non-conforming unmanned aerial systems,'' in \emph{2020 AIAA/IEEE 39th
  Digital Avionics Systems Conference (DASC)}, 2020, pp. 1--10.

\bibitem{IFGreen18}
I.~Fyfe-Green, ``A study of the perceptions of air safety and mid-air collision
  prevention during regulatory change,'' Ph.D. dissertation, Business
  Administration of the University of Portsmouth, 2018.

\bibitem{ca92069d33d24c49b280c575f9b5c8ef}
C.~Luo, S.~McClean, G.~Parr, L.~Teacy, and R.~{De Nardi},
  ``\BIBforeignlanguage{English}{Uav position estimation and collision
  avoidance using the extended kalman filter},''
  \emph{\BIBforeignlanguage{English}{IEEE Trans Veh Technol}}, vol.~62, no.~6,
  pp. 2749--62, 2013.

\bibitem{GWelch06}
G.~Welch and G.~Bishop, ``An introduction to the kalman filter,'' \emph{Proc.
  Siggraph Course}, vol.~8, 01 2006.

\bibitem{5937290}
R.~W. Osborne, Y.~Bar-Shalom, P.~Willett, and G.~Baker, ``Design of an adaptive
  passive collision warning system for uavs,'' \emph{IEEE Transactions on
  Aerospace and Electronic Systems}, vol.~47, no.~3, pp. 2169--89, 2011.

\bibitem{9637501}
K.~Kuru, ``Conceptualisation of human-on-the-loop haptic teleoperation with
  fully autonomous self-driving vehicles in the urban environment,'' \emph{IEEE
  Open J. Intell. Transp. Syst.}, vol.~2, pp. 448--9, 2021.

\bibitem{UKCAA23}
\BIBentryALTinterwordspacing
UKCAA, ``Mid air collision: Our safety plan,'' 2023. [Online]. Available:
  \url{https://www.caa.co.uk/safety-initiatives-and-resources/how-we-regulate/safety-plan/mitigating-key-safety-risks/mid-air-collision/}
\BIBentrySTDinterwordspacing

\bibitem{9498989}
J.~Tang, G.~Liu, and Q.~Pan, ``A review on representative swarm intelligence
  algorithms for solving optimization problems: Applications and trends,''
  \emph{IEEE/CAA J. Autom. Sin.}, vol.~8, no.~10, pp. 1627--43, 2021.

\bibitem{Kkuru24}
\BIBentryALTinterwordspacing
K.~Kuru, S.~Worthington, D.~Ansell, J.~M. Pinder, A.~Sujit, B.~Jon~Watkinson,
  K.~Vinning, L.~Moore, C.~Gilbert, D.~Jones, and C.~L. Tinker-Mill,
  ``Aitl-wing-hitl: Telemanipulation of autonomous drones using digital twins
  of aerial traffic interfaced with wing,'' \emph{IEEE Access}, 2023,
  submitted. [Online]. Available: \url{https://clok.uclan.ac.uk/47765/}
\BIBentrySTDinterwordspacing

\bibitem{YBSebbane15}
Y.~B. Sebbane, \emph{Smart Autonomous Aircraft: Flight Control and Planning for
  UAV Hardcover}.\hskip 1em plus 0.5em minus 0.4em\relax FL, USA: CRC Press,
  2016.

\bibitem{8494201}
P.~{Chen} and C.~{Lee}, ``Uavnet: An efficient obstacel detection model for uav
  with autonomous flight,'' in \emph{2018 International Conference on
  Intelligent Autonomous Systems}, 2018, pp. 217--220.

\bibitem{doi:10.2514/6.2012-2455}
C.~Sabo and K.~Cohen, \emph{Dynamic Allocation of Unmanned Aerial Vehicles with
  Communication Constraints}.\hskip 1em plus 0.5em minus 0.4em\relax AIAA,
  2012.

\bibitem{9181515}
Y.~Mao, M.~Chen, X.~Wei, and B.~Chen, ``Obstacle recognition and avoidance for
  uavs under resource-constrained environments,'' \emph{IEEE Access}, vol.~8,
  pp. 169\,408--169\,422, 2020.

\bibitem{4967999}
D.~J. {Pack}, P.~{DeLima}, G.~J. {Toussaint}, and G.~{York}, ``Cooperative
  control of uavs for localization of intermittently emitting mobile targets,''
  \emph{IEEE Transactions on Systems, Man, and Cybernetics, Part B
  (Cybernetics)}, vol.~39, no.~4, pp. 959--970, 2009.

\bibitem{7423671}
M.~{Gharibi}, R.~{Boutaba}, and S.~L. {Waslander}, ``Internet of drones,''
  \emph{IEEE Access}, vol.~4, pp. 1148--1162, 2016.

\bibitem{9061133}
E.~Vinogradov, F.~Minucci, and S.~Pollin, ``Wireless communication for safe
  uavs: From long-range deconfliction to short-range collision avoidance,''
  \emph{IEEE Veh Technol Mag}, vol.~15, no.~2, pp. 88--95, 2020.

\end{thebibliography}

\begin{IEEEbiography}[{\includegraphics[width=1in,height=1.25in,clip,keepaspectratio]{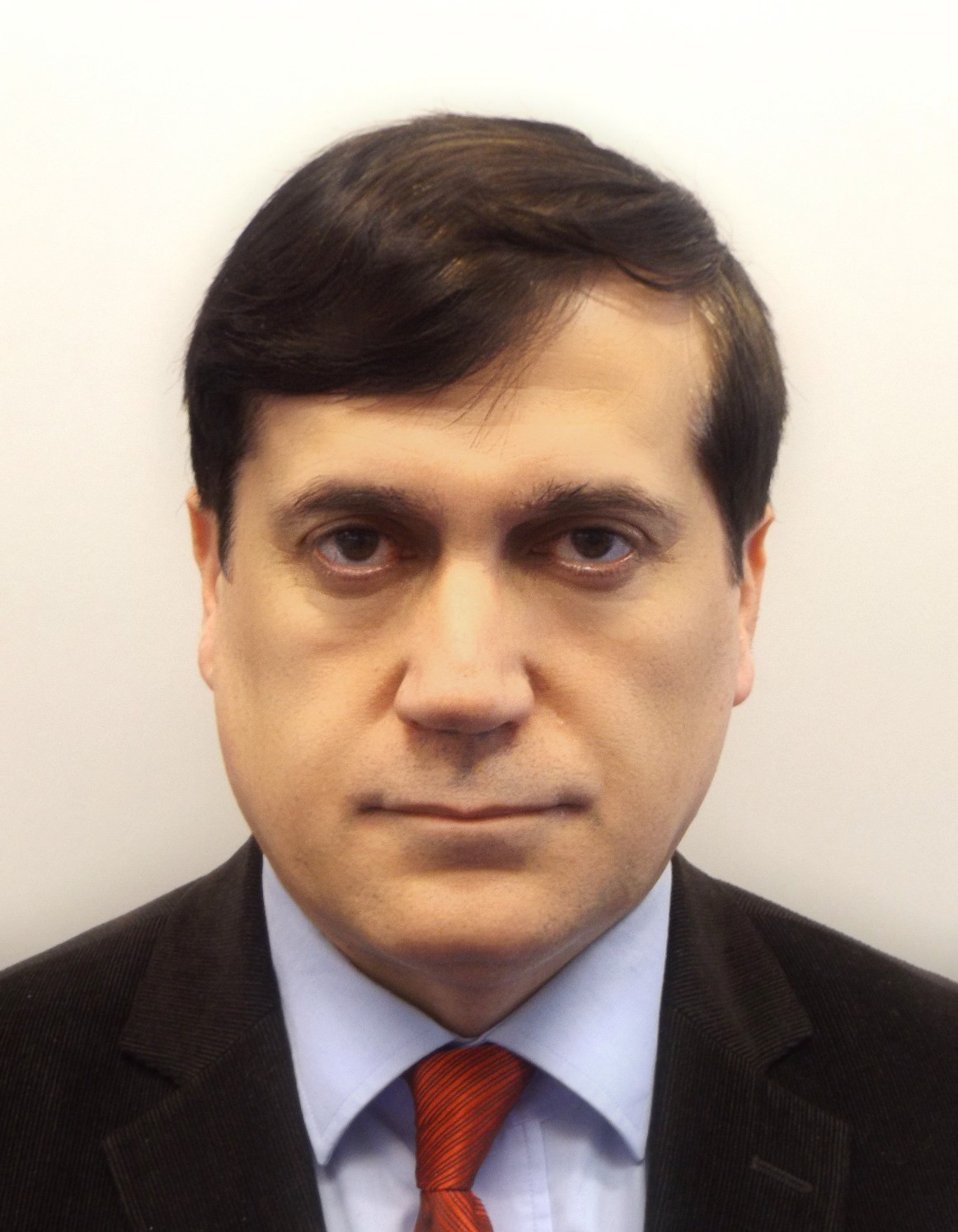}}]{Kaya Kuru} received the B.Sc. degree from National Defense University (Turkish Military Academy), the major/ADP degree in computer engineering from Middle East Technical University (METU), the M.B.A. degree from Selcuk University, the M.Sc. and Ph.D. degrees in computer science from METU. He completed his postdoctoral studies with the School of Electronics and Computer Science/Informatics, University of Southampton, UK. 
He is a Software Engineer and currently an Associate Professor of Computer-Information Systems Engineering.
He has recently engaged in the implementation of numerous AI-based real-world systems within various funded projects. 
His research interests include the development of geo-distributed autonomous intelligent systems using FL, ML, DL, and DRL with CPSs.
\end{IEEEbiography}

\begin{IEEEbiography}[{\includegraphics[width=1in,height=1.25in,clip,keepaspectratio]{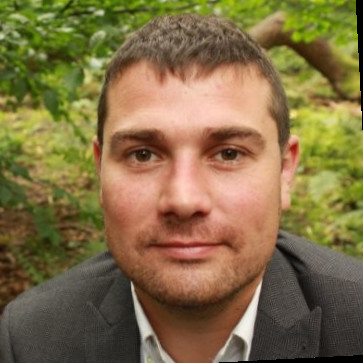}}]{John Michael Pinder} received a B.Sc. degree in AI from the University of Sunderland, M.Sc. in Robotics and Automation and Ph.D. degrees in Autonomous Systems from the University of Salford. He is Data Scientist and has over 20 years experience applying technology to solve practical problems and generate value for customers. Previously, Dr. Pinder worked for a Robotics \& Automation control software company, helping medium-to-large enterprises across a wide range of industrial sectors to develop bespoke automated machinery. 
\end{IEEEbiography}

\begin{IEEEbiography}[{\includegraphics[width=1in,height=1.25in,clip,keepaspectratio]{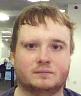}}]{Benjamin Jon Watkinson} has the M.Sc. in Robotics Engineering from the University of Central Lancashire. He has 9 years experience at UCLan working with NASA, BAE Systems and several SMEs supporting product development activities and research. He is currently Hardware/Flight Test integration lead. His research interests include the development of bespoke robotics systems and drone systems to solve real-world problems. 
\end{IEEEbiography}


\begin{IEEEbiography}[{\includegraphics[width=1in,height=1.25in,clip,keepaspectratio]{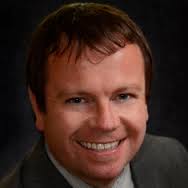}}]{Darren Ansell} received the B.SC. from the University of Manchester Institute of Science and Technology in electrical and electronic engineer and Ph.D. from Cranfield University in antenna optimisation using evolutionary algorithms. He is the engineering lead for Space and Aerospace and Professor in Aerospace Engineering. He previously worked in industry at BAE Systems in management roles, specialising in Mission Systems and Autonomy. 
\end{IEEEbiography}

\begin{IEEEbiography}[{\includegraphics[width=1in,height=1.25in,clip,keepaspectratio]{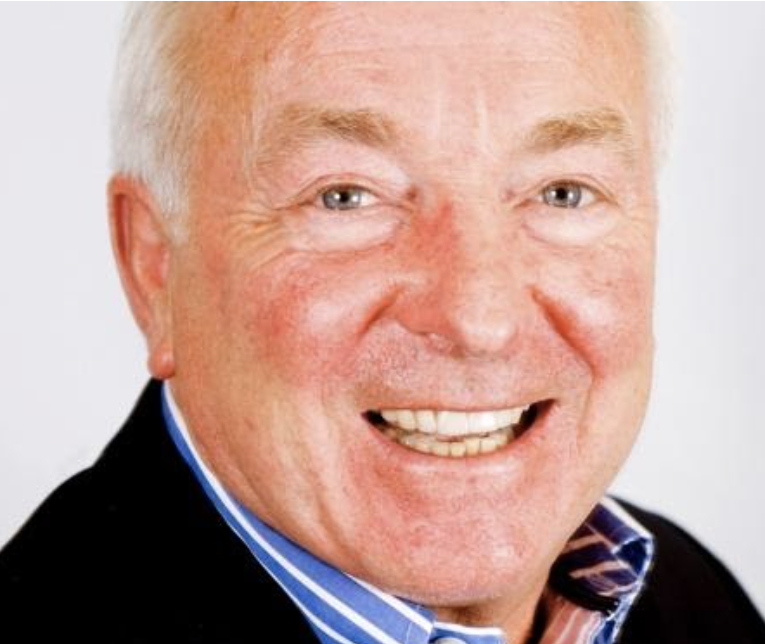}}]{KEITH VINNING} OBE is the project lead in flight testing and development support and is a GA pilot with 50 years of flight experience. Keith received a BSc Honours degree in Electrical Engineering and Telecommunications from Aston University in 1981. He gained 49 years of experience in telecommunications product and business development, holding senior roles within BT and Fujitsu in the UK and Europe. Joined PilotAware in 2015 and was awarded an OBE in 2020 for services to Aviation Safety based on the work done with PilotAware.	
\end{IEEEbiography}

\begin{IEEEbiography}[{\includegraphics[width=1in,height=1.25in,clip,keepaspectratio]{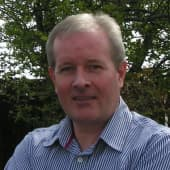}}]{Lee Moore} is CEO of PilotAware. He is software and systems lead and Pilot. 35 years in semiconductor and electronic design automation, digital ASIC design, processor design, software design tools. Experienced in both multinational and start-up companies GEC Telecommunications, NEC Semiconductors, Synopsys Inc, Cadence Design Systems Inc.
\end{IEEEbiography}
\begin{IEEEbiography}[{\includegraphics[width=1in,height=1.25in,clip,keepaspectratio]{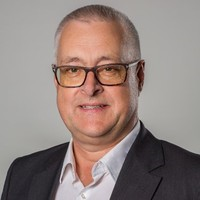}}]{Chris~Gilbert} received the B.Sc. degree from University of Bristol in Physics. He is Network Lead Pilot, General Manager, Motorola Networks EMEA, CEO of wireless network start-ups in Silicon Valley and the UK, Director of a chip company and two security companies and PilotAware.
\end{IEEEbiography}
\begin{IEEEbiography}[{\includegraphics[width=1in,height=1.25in,clip,keepaspectratio]{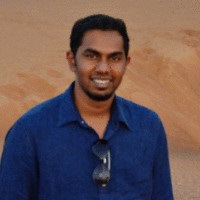}}]{Aadithya Sujit} received a B.Sc in Mechanical Engineering from Visvesvaraya Technological University and a M.Sc in Aerospace engineering from Linkoping university, Sweden. He specializes in aerospace vehicle design \& development and is currently supporting industries with R\&D activities. 
	He has previously worked at TU Delft in Netherlands developing bio inspired UAVs, prior to which he worked at National Aerospace Laboratories in India as a UAV Project Engineer.
\end{IEEEbiography}
\begin{IEEEbiography}[{\includegraphics[width=1in,height=1.25in,clip,keepaspectratio]{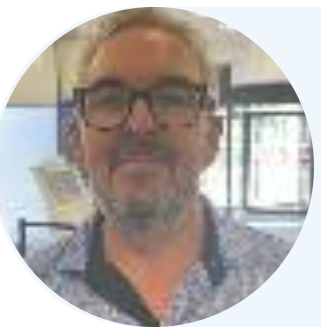}}]{David Jones} received the A' Levels degree from St. Austin's Academy in Economics, Maths, Physics. He is a professional freelance videographer and chief drone pilot providing services in film making and surveying with PfCO. His background in outdoor sports helps a lot when it comes to working with adventure companies that want to create amazing videos to help their own advertising and marketing. 
\end{IEEEbiography}		

	\EOD

\end{document}